# Large-Scale Neuromorphic Spiking Array Processors: A quest to mimic the brain

Chetan Singh Thakur[1*], Jamal Molin[2], Gert Cauwenberghs[3], Giacomo Indiveri[4], Kundan Kumar[1], Ning Qiao[4], Johannes Schemmel[5], Runchun Wang[6], Elisabetta Chicca[7], Jennifer Olson Hasler[8], Jae-sun Seo[9], Shimeng Yu[9], Yu Cao[9], André van Schaik[6], Ralph Etienne-Cummings[2]

[1]Department of Electronic Systems Engineering, Indian Institute of Science, Bangalore, India

[2]Department of Electrical and Computer Engineering Johns Hopkins University Baltimore, MD, USA

[3]Department of Bioengineering and Institute for Neural Computation, University of California, San Diego, La Jolla, CA, USA

[4]Institute of Neuroinformatics, University of Zurich and ETH Zurich, Zurich, Switzerland

[5]Kirchhoff Institute for Physics, University of Heidelberg, Heidelberg, Germany

[6]The MARCS Institute, Western Sydney University, Kingswood, NSW 2751, Australia

[7]Cognitive Interaction Technology – Center of Excellence, Bielefeld University, Bielefeld, Germany

[8]School of Electrical and Computer Engineering, Georgia Institute of Technology, Atlanta, GA, USA

[9]School of Electrical, Computer and Engineering, Arizona State University, Tempe, AZ, USA

*Correspondence: Chetan Singh Thakur, csthakur@iisc.ac.in

Abstract

Neuromorphic engineering (NE) encompasses a diverse range of approaches to information processing that are inspired by neurobiological systems, and this feature distinguishes neuromorphic systems from conventional computing systems. The brain has evolved over billions of years to solve difficult engineering problems by using efficient, parallel, low-power computation. The goal of NE is to design systems capable of brain-like computation. Numerous large-scale neuromorphic projects have emerged recently. This interdisciplinary field was listed among the top 10 technology breakthroughs of 2014 by the MIT Technology Review and among the top 10 emerging technologies of 2015 by the World Economic Forum. NE has two-way goals: one, a scientific goal to understand the computational properties of biological neural systems by using models implemented in integrated circuits (ICs); second, an engineering goal to exploit the known properties of biological systems to design and implement efficient devices for engineering applications. Building hardware neural emulators can be extremely useful for simulating large-scale neural models to explain how intelligent behavior arises in the brain. The principle advantages of neuromorphic emulators are that they are highly energy efficient, parallel and distributed, and require a small silicon area. Thus, compared to conventional CPUs, these neuromorphic emulators are beneficial in many engineering applications such as for the porting of deep learning algorithms for various recognitions tasks. In this review article, we describe some of the most significant neuromorphic spiking emulators, compare the different architectures and approaches used by them, illustrate their advantages and drawbacks, and highlight the capabilities that each can deliver to neural modelers. This article focuses on the discussion of large-scale emulators and is a continuation of a previous review of various neural and synapse circuits [1]. We also explore applications where these emulators have been used and discuss some of their promising future applications.

## 1 Introduction

Building a vast digital simulation of the brain could transform neuroscience and medicine and reveal new ways of making more powerful computers [2]. The human brain is by far the most computationally complex, efficient, and robust computing system operating under low-power and small-size constraints. It utilized over 100 billion neurons and 100 trillion synapses for achieving these specifications. Even the largest supercomputing systems to date are not capable of obtaining real-time performance when running simulations large enough to accommodate multiple cortical areas, yet detailed enough to include distinct cellular properties. For example, for mouse-scale ($2.5 \times 10^6$ neurons) cortical simulations, a personal computer uses 40,000 times more power but runs 9000 times slower than a mouse brain [3]. The simulation of a human-scale cortical model ($2 \times 10^{10}$ neurons), which is the goal of the Human Brain Project, is projected to require an exascale supercomputer (1018 flops) and as much power as a quarter-million households (0.5 GW).

The electronics industry is seeking solutions that enable computers to handle the enormous increase in data processing requirements. Neuromorphic computing is an alternative solution that is inspired by the computational capabilities of the brain. The observation that the brain operates on analog principles of the physics of neural computation that are fundamentally different from digital principles in traditional computing has initiated investigations in the field of neuromorphic engineering [4]. Silicon neurons (SiNs) are hybrid analog/digital very-large-scale integrated (VLSI) circuits that emulate the electrophysiological behavior of real neurons and synapses. Neural networks using SiNs can be emulated directly in hardware rather than being limited to simulations on a general-purpose computer. Such hardware emulations are much more energy efficient than computer simulations, and thus suitable for real-time large-scale neural emulations. The hardware emulations operate in real-time and the speed of the network can be independent of the number of neurons or their coupling.

There has been growing interest in neuromorphic processors to perform real-time pattern recognition tasks, such as object recognition and classification, owing to the low energy and silicon area requirements of these systems [5][6]. These large systems will find application in the next generation of technologies including autonomous cars, drones, and brain-machine interfaces. The neuromorphic chip market is expected to grow exponentially owing to an increasing demand for artificial intelligence and machine learning systems and the need for better-performing ICs and new ways of computation as Moore's law is pushed to its limit [7].

The biological brains of cognitively sophisticated species have evolved to organize their neural sensory information processing with computing machinery that are highly parallel and redundant, yielding great precision and efficiency in pattern recognition and association despite operating with intrinsically sluggish, noisy, and unreliable individual neural and synaptic components. Brain-inspired neuromorphic processors show great potential for building compact natural signal processing systems, pattern recognition engines, and real-time behaving autonomous agents [8]–[10]. Profiting from their massively parallel computing substrate [9] and co-localized memory and computation features, these hardware devices have the potential to solve the von Neumann memory bottleneck problem [11] and to reduce power consumption by several orders of magnitude. Compared to pure digital solutions, mixed-signal neuromorphic processors offer additional advantages in terms of lower silicon area usage, lower power consumption, reduced bandwidth requirements, and additional computational complexity.

Several neuromorphic systems are already being used commercially. For example, Synaptics Inc. develops touchpad and biometric technologies for portable devices, Foveon Inc. develops Complementary Metal Oxide-Semiconductor (CMOS) color imagers [12], and Chronocam Inc. builds asynchronous time-based image sensors based on the work in [13]. Another product, an artificial retina, is being used in the Logitech Marble trackball, which optically measures the rotation of a ball to move the cursor on a computer screen [14]. The dynamic vision sensor (DVS) by iniLabs Ltd. is another successful neuromorphic product [15].

In this work, we describe a wide range of neural processors based on different neural design strategies and synapses that range from current-mode, sub-threshold to voltage-mode, switched-capacitor designs. Moreover, we will discuss the advantages and strengths of each system and their potential applications.

## 2      Integrate-and-Fire Array Transceiver (IFAT)

The Integrate-and-Fire Array Tranceiver (IFAT) is a mixed-mode VLSI-based neural array with reconfigurable, weighted synapses/connectivity. In its original design, it is comprised of the array of mixed-mode VLSI neurons, an LUT (look-up table), and AER (Address Event Representation) architecture. The AER architecture is used for the receiver and transmitter. The AER communication protocol is an event-based, asynchronous protocol. Addresses are inputs to the chip (address-events). The addresses represent the neuron receiving the input event/spike. When a neuron outputs an event, it outputs the address (output address-event) of the neuron emitting the event/spike. The LUT holds the information on how the network is connected. It consists of the corresponding destination address(es) (postsynaptic events) for each incoming address-event (presynaptic event). For each connection, there is a corresponding weight signifying the strength of the postsynaptic event. The larger the weight, the more charge integrated onto the membrane capacitance of the destination neuron receiving the postsynaptic event. There is also a polarity bit corresponding to each synapse signifying an inhibitory or excitatory postsynaptic event. The original design of the IFAT utilized probabilistic synapses [16]. The weights were represented by a probability. As events were received, the probability of the neuron receiving the event was represented as the weight. The neuron circuit used was essentially a membrane capacitor coupled to a comparator and a synapse implemented as a transmission gate and charge pump. The next generation of the IFAT used conductance based synapses [17]. Instead of representing weights as probabilities, a switch-cap circuit was used. The weight then represented the synapse capacitance, and therefore, was proportional to the amount of charge integrated onto the membrane capacitance.

In sections 2.1 and 2.2, two novel derivations of the IFAT will be depicted: MNIFAT and HiAER IFAT.

## 2.1 __MNIFAT:__ (Mihalas-Niebur and Integrate-and-Fire Array Transceiver)

This section describes novel integrate-and-fire array transceiver (IFAT) neural array (MNIFAT), which consists of 2040 Mihalas-Niebur (M-N) neurons. The M-N neuron circuit design used in this array was shown to produce nine prominent spiking behaviors using an adaptive threshold. Each of these M-N neurons were designed to have the capability to operate as two independent integrate-and-fire (I&F) neurons. This resulted in 2040 M-N neurons and 4080 leaky I&F neurons. This neural array was implemented in 0.5 μm CMOS technology with a 5 V nominal power supply voltage [15]. Each I&F consumes an area of 1495 μm$^2$, while the neural array dissipates an average of 360 pJ of energy per synaptic event at 5 V. This novel neural array design consumes considerably less power and area per neuron than other neural arrays designed in CMOS technology of comparable feature size. Furthermore, the nature of the design allows for more controlled mismatch between neurons.

### 2.1.1 Neural array design

The complete block diagram of the neuron array chip is shown in Figure 1. It was implemented with the aim to maximize the neuron array density, minimize power consumption, and reduce mismatch due to process variation. This is achieved by utilizing a single membrane synapse (switch-capacitor circuit) and soma (comparator) shared by all neurons in the array. The connections between the neurons is reconfigurable via an off-chip LUT. Presynaptic events are sent first through the LUT where the destination addresses and synaptic strengths are stored. Post-synaptic events are then sent to the chip. These events are sent as address-events (AE) along a shared address bus decoded by the row decoder and column decoder on-chip. The incoming address corresponds to a single neuron in the array.

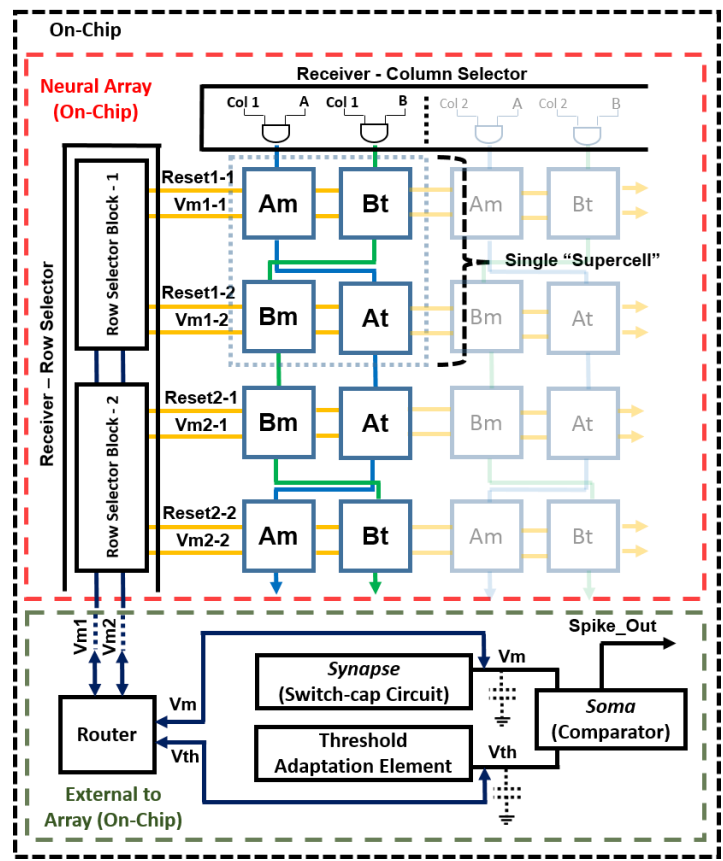

Figure 1. Mihalas-Niebur neural array design.

The neuron array is made up of supercells, each containing four cells, labeled Am, At, Bm, and Bt. Each supercell contains two M-N neurons, one using Am and At cells, and the second using Bm and Bt cells. Each of these M-N neurons can also operate as two independent leaky (I&F) neurons, resulting in a total of four leaky I&F neurons (Am, At, Bm, and Bt). Incoming AE selects the supercell in the array and consists of two additional bits for selecting one of the two M-N neurons (A or B) within the supercell, or one of the four cells when operating as I&F neurons. Finally, the voltage across the storage capacitance for both the membrane cell and threshold cell is buffered to the processor via the router (Vm1-X and Vm2-X, where X is the row selected). The router is used for selecting which voltage (from the membrane cell or threshold cell) is buffered to the processor as the membrane voltage and/or threshold voltage, depending on the mode selected (M-N mode or I&F mode). This router is necessary for allowing the voltage from the threshold (At or Bt) cell to be used as the membrane voltage when in I&F mode. After the selected neuron cell(s) buffer their stored voltage to the external capacitances Cm and Ct, the synaptic event is applied and the new voltage is buffered back to the same selected cells that received the event. The synapse and threshold adaptation elements execute the neuron dynamics as events are received. If the membrane voltage exceeds the threshold voltage, there is a single comparator (soma) that outputs a logic high (event).

An output arbiter/transmitter is not necessary in our design considering that a neuron only fires when it receives an event. The single output signal always corresponds to the neuron that receives the incoming event. Having a single comparator not only reduces power consumption but also reduces the required number of pads for digital output. In this design, the speed is compromised (for low-power and low-area) due to the time necessary to read and write to and from the neuron. However, a maximum input event rate of ~1 MHz can still be achieved for proper operation.

### 2.1.2 Mihalas-Niebur(M-N) neuron model and circuit implementation

Each cell pair (Am/At and Bm/Bt) in this neural array models the M-N neuron dynamics [18]. In its original form, it uses linear differential equations and parameters with biological facsimiles. It consists of an adaptive threshold and was shown to be capable of modeling all of the biologically relevant neuron behaviors. It uses three differential equations modeling the internal currents (Eq. 1), membrane voltage (Eq. 2), and adaptive threshold voltage (Eq. 3) dynamics:

$$I_j'(t) = -k_j I_j(t); j = 1, \dots, N \tag{1}$$

$$V_m'(t) = \frac{1}{C}(I_{ext} + \sum_j I_{j(t)} - G(V)m(t) - E_L)) \tag{2}$$

$$\theta'(t) = a(V_m(t) - E_L) - b(\theta(t) - \theta_\infty) \tag{3}$$

$I_j$ represents the internal, spike-induced currents. $I_{ext}$ is the external input to the neuron. $V_m$ is the membrane voltage. $\theta$ is the adaptive threshold. $k_j, G, a, b, E_L, and \theta_\infty$ are free parameters. An output spike occurs when the membrane voltage, $V_m$, exceeds the threshold voltage, $\theta$. Following an output spike, the following update rules (Eq. 4, Eq. 5, and Eq. 6) are applied for the internal currents, membrane voltage, and adaptive threshold, respectively.

$$I_j(t) \leftarrow R_j \times I_j(t) + A_j \tag{4}$$

$$V_m(t) \leftarrow V_r \tag{5}$$

$$\theta(t) \leftarrow max(\theta_r, \theta(t)) \tag{6}$$

$R_j$, $A_j$, $V_r$, $\theta_r$ are free parameters. $V_r$ and $\theta_r$ are the membrane reset and adaptive threshold reset voltages, respectively. This original M-N was capable of producing all 20 of the prominent spiking behaviors.

For the circuit implementation of this M-N model, a few modifications to the original model were made. First, internal induced-spike currents were omitted, and second, the reset voltage was set equal to the resting potential. These modifications were made at the expense of the generality of the model. However, this modified M-N model is still capable of implementing Eq. (9) of the biologically relevant spiking behaviors. The modified differential equations for CMOS implementation are as follows:

$$V'_m(t) = \frac{g_l^m}{C_m}(V_r - V_m(t)) \tag{7}$$

$$\theta'(t) = \frac{g_l^t}{C_t}(\theta_r - \theta(t)) \tag{8}$$

$$V_m(t+1) = V_m(t) + \frac{C_s^m}{C_m}(E_m - V_m(t)) \tag{9}$$

$$\theta(t+1) = \theta(t) + \frac{C_s^t}{C_t}(V_m(t) - V_r) \tag{10}$$

and,

$$g_l^m = \frac{1}{r_l^m} = f_l^m C_l \tag{11}$$

$$g_l^t = \frac{1}{r_l^t} = f_l^t C_l \tag{12}$$

As previously discussed, this modification allows the use of two neuron cells to work together to model a single M-N neuron. Eq. (9) and (10) model the change in membrane potential ($V_m$) and threshold potential ($\theta$) at each time step as the neuron receives an input. $C_s^m$ and $C_s^t$ are the switch-capacitor capacitance depicting the synapse conductance or threshold adaptation conductance, respectively. $C_m$ and $C_t$ are the storage capacitance for the membrane and threshold cells, respectively. $E_m$ is the synaptic driving potential. Eq. (7) and (8) model the leakage dynamics, independent of synaptic connections. $g_l^m$ and $g_l^t$ are the leakage conductances for the membrane and threshold and are dependent on the clock frequency, $f_l^m$ and $f_l^t$. The update rules for this modified M-N neuron model are as follows:

$$V_m(t) \leftarrow V_r \tag{13}$$

$$\theta(t) \leftarrow \{if\ \theta(t) > V_m(t), \theta(t) if\ \theta(t) \leq V_m(t), \theta_r \tag{14}$$

### 2.1.3 Neuron cell circuit

The neuron cell circuit is shown in Figure 2. The PMOS transistor, P1, is the storage capacitance (~440 fF), $C_m$ or $C_t$, (depending on whether the cell is being used to model the membrane or threshold dynamics) implemented as a MOS capacitor with its source and drain tied to Vdd. Transistors N1 and N2 model the leakage (Eq. 7 and 8) via a switch capacitor circuit with Phi1 and Phi2 pulses at a rate of $f_l^{m,t}$ (also $C_l \ll C_m$). Transistors N3 and N4 allow for resetting the neuron when selected (ColSel = "1"). Transistor N5 forms a source-follower when coupled with a globally shared variable resistance located in the processor of the neural array. It is implemented as an NMOS transistor with a voltage bias ($V_b$). In read mode (RW = "0"), switch S2 is closed such that the voltage across the storage capacitance is buffered to an equivalent capacitance coupled to the synapse and/or threshold adaptation element. In write mode (RW = "1"), switch S3 is closed such that the new voltage from the synapse/threshold elements (after an event is received) is buffered to the storage capacitance.

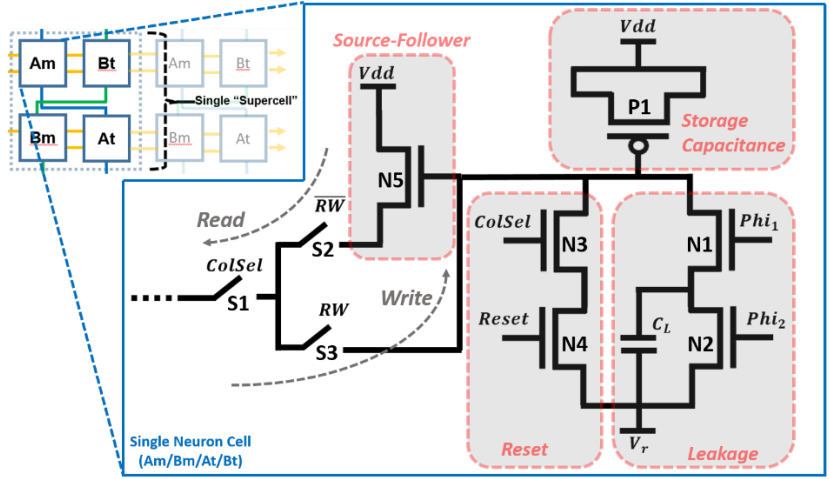

Figure 2. Single neuron cell design.

### 2.1.4 Synapse and threshold adaptation circuits

The schematic for modeling the neuron dynamics can be seen in Figure 3. When a neuron receives an event, RW = "0", and the neuron's cell is selected and its stored membrane voltage is buffered to the capacitance Cm. In the same manner, if in M-N mode, the threshold voltage is buffered to Ct. The Phi1$_{SC}$ and Phi2$_{SC}$ pulses are then applied (off-chip), adding (excitatory event) or removing (inhibitory event) charge to Cm via the synapse using a switch-capacitor circuit. A second, identical switch-capacitor circuit is used for implementing the threshold adaptation dynamics. As a neuron receives events, the same Phi1$_{SC}$ and Phi2$_{SC}$ pulses are applied to the threshold adaptation switch-capacitor circuit which adds or removes charge to Ct. The new voltage is then buffered (RW = "1") back to the neuron cells for storing the new membrane voltage (as well as the threshold voltage if in M-N mode). When using each neuron independently as leaky I&F neurons, the threshold adaptive element is bypassed and an externally applied fixed threshold voltage is used. A charge-based subtractor is used in the threshold adaptation circuit for computing $V_{th} + (V_m - V_r)$ in modelling Eq. (10). This subtraction output is the driving potential for the threshold switch-capacitor circuit. An externally applied voltage, $E_m$, is the synaptic driving potential for the membrane synapse and is used for modeling Eq. (9). Finally, the comparator outputs an event when the membrane voltage exceeds the threshold voltage. An external reset signal for both the neuron cell modeling the membrane voltage and cell modeling the threshold voltage is activated for the selected neuron (via Reset1-X and Reset2-X) when a spike is outputted.

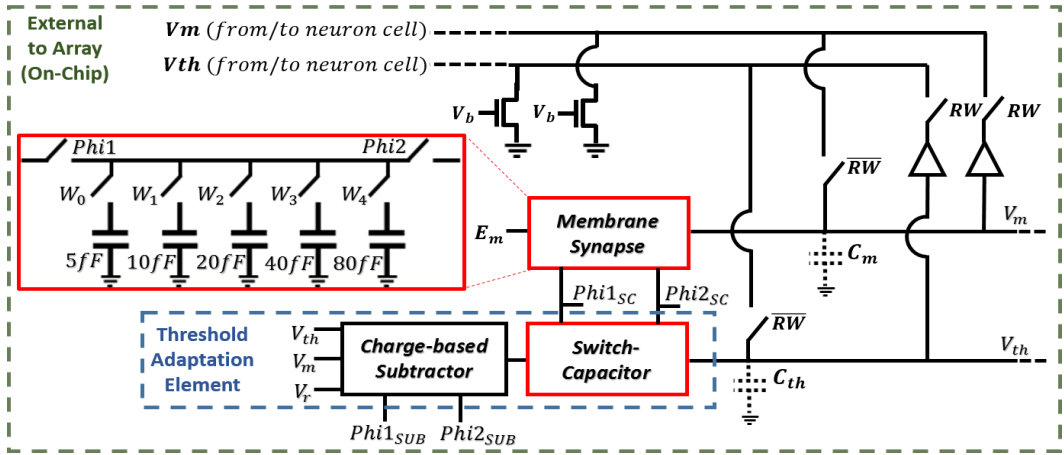

Figure 3. Block diagram of the processor including the synapse and threshold adaptation circuits.

### 2.1.5 Results

A single neuron cell in this array has dimensions of 41.7 μm × 35.84 μm. It consumes only 62.3% of the area consumed by a single neuron cell in [19], also designed in a 0.5 μm process. While we achieve 668.9 I&F neurons/mm$^2$, [19] achieves only 416.7 neurons/mm$^2$ and [20] achieves only 387.1 neurons/mm$^2$. The number of neurons/mm$^2$ can be further increased by optimizing the layout of the neuron cell and implementing in smaller feature-size technology.

We also analyzed the mismatch (due to process variations) across the neuron array, we observed the output event to input event ratio for a fixed synaptic weight and an input event rate of 1 MHz for each neuron in the array. With this fixed synaptic weight, the 2040 M-N neurons have a mean output to input event ratio of 0.0208 ±1.22e-5. In the second mode of operation, the 4080 I&F neurons have a mean output to input event ratio of 0.0222 ±5.57e-5. For fair comparison, we compare with the results from a similar experiment performed in the 0.5 μm conductance-based IFAT in [19] (Table 1). Our design shows significantly less deviation. Small amounts of mismatch can be taken advantage of in applications that require stochasticity. However, for those spike-based applications that do not benefit from mismatch, in this neural array, it is more controlled. This again is a result of utilizing a single, shared synapse, comparator, and threshold adaptive element for all neurons in the array. The mismatch between neurons is only due to the devices within the neuron cell itself.

Table 1. Array characterization showing output events per input event

| Neural Array | Mean Ratio (μ) | Standard Deviation (σ) | # of Neurons |
|---|---|---|---|
| IFAT | 0.0222 | ±5.57e-5 | 4080 |
| Vogelstein et. al [19] | 0.0210 | ±1.70e-3 | 2400 |

Another design goal was to minimize power consumption. At an input event rate of 1 MHz, the average power consumption was 360 μW at 5.0 V power supply. A better representation of the power consumption is energy per incoming event. From these measurements, this chip consumes 360 pJ of energy per synaptic event. A comparison with other state-of-the-art neural array chips can be seen in Table 2. Compared to those chips designed in 500 nm [19] and 800 nm [21] technology, a significant reduction in energy per synaptic event was seen. Due to complications in the circuit board, the low-voltage operation could not be measured. However, the proper operation at 1.0 V (at slower speeds) was validated using simulations. Assuming dynamic energy scales with $V^2$ (capacitance remains the same), the energy per synaptic event was estimated as ~14.4 pJ at 1.0 V. These results are promising, as these specifications will be even further optimized by designing in smaller feature-size technology. Table 2 summarizes the specifications achieved from this chip.

Table 2. Measured (*estimated) chip results

| Process | Vdd Supply | (I&F) Neuron Density | Neuron Area | Energy/Event |
|---------|-----------|---------------------|-------------|--------------|
| $500nm$ | $5.0V$ | $669\ neurons/mm^2$ | $1495\mu m^2$ | $360pJ$ |
| $55nm^*$ | $1.2V$ | $55298\ neurons/mm^2$ | $18.1\mu m^2$ | $20.7pJ$ |

## 2.2 HiAER-IFAT: Hierarchical Address-Event Routing (HiAER) Integrate-and-Fire Array Transceivers (IFAT)

Hierarchical address-event routing Integrate-and-Fire Array Transceiver (HiAER-IFAT) provides a multiscale tree-based extension of AER synaptic routing for dynamically reconfigurable long-range synaptic connectivity in neuromorphic computing systems, developed in the lab of Gert Cauwenberghs at University of California San Diego. A major challenge in scaling up neuromorphic computing to the dimensions and complexity of the human brain, a necessary endeavor towards bio-inspired general artificial intelligence approaching human-level natural intelligence, is to accommodate massive flexible long-range synaptic connectivity between neurons across the network in highly efficient and scalable manner. Meeting this challenge calls for a multi-scale system architecture, akin to the organization of grey and white matter distinctly serving local compute and global communication functions in the biological brain, that combines highly efficient dense local synaptic connectivity with highly flexible reconfigurable sparse long-range connectivity.

Efforts towards this objective for large-scale emulation of neocortical vision have resulted in event-driven spiking neural arrays with dynamically reconfigurable synaptic connections in a multi-scale hierarchy of compute and communication nodes abstracting such grey and white matter organization in the visual cortex (Figure 4) [22]–[26]. Hierarchical address-event routing (HiAER) offers scalable long-range neural event communication tailored to locally dense and globally sparse synaptic connectivity [24], [27], while integrate-and-fire array transceiver (IFAT) CMOS neural arrays with up to 65k neurons integrated on a single chip [28]–[30] offer low-power implementation of continuous-time analog membrane dynamics at energy levels down to 22 pJ/spike [23].

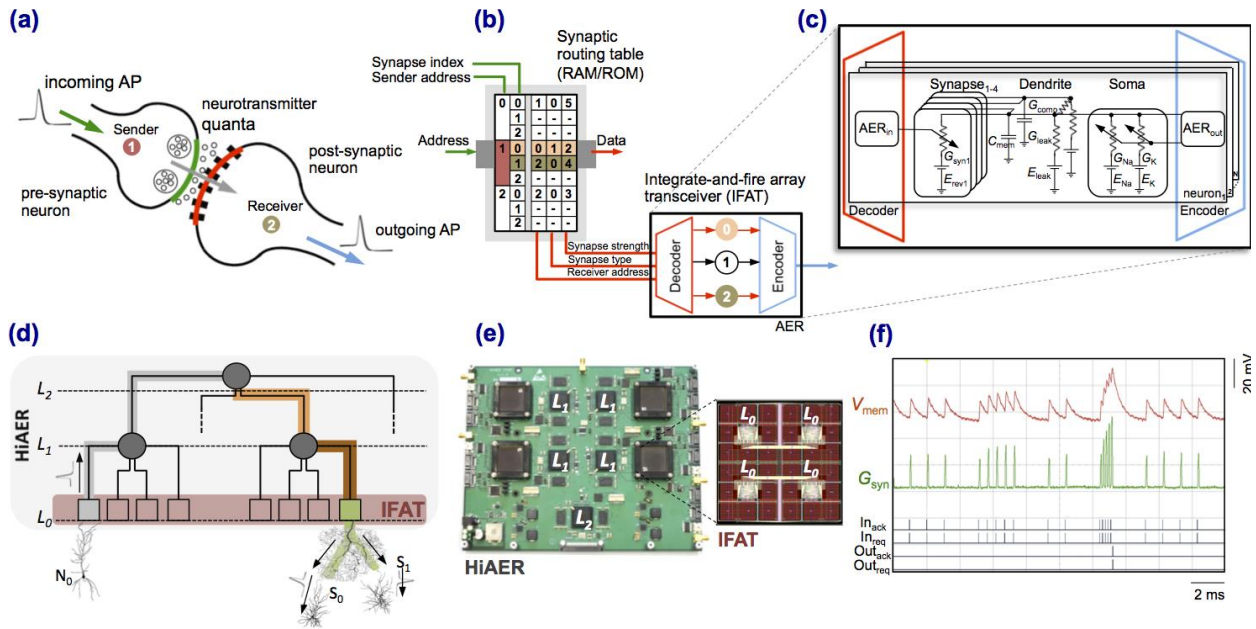

Figure 4. Hierarchical Address-Event Routing (HiAER) Integrate-and-Fire Array Transceiver (IFAT) for scalable and reconfigurable neuromorphic neocortical processing [24], [31]. (a) Biophysical model of neural and synaptic dynamics. (b) Dynamically reconfigurable synaptic connectivity is implemented across IFAT arrays of addressable neurons by routing neural spike events locally through DRAM synaptic routing tables [28], [29]. (c) Each neural cell models conductance-based membrane dynamics in proximal and distal compartments for synaptic input with programmable axonal delay, conductance, and reversal potential [23], [30]. (d) Multiscale global connectivity through a hierarchical network of HiAER routing nodes [27]. (e) HiAER-IFAT board with 4 IFAT custom silicon microchips, serving 256k neurons and 256M synapses, and spanning 3 HiAER levels (L0-L2) in connectivity hierarchy [24]. (f) The IFAT neural array multiplexes and integrates (top traces) incoming spike synaptic events to produce outgoing spike neural events (bottom traces) [30]. The most recent IFAT microchip-measured energy consumption is 22 pJ per spike event [23], several orders of magnitude more efficient than emulation on CPU/GPU platforms.

The energy efficiency of such large-scale neuromorphic computing systems is limited primarily by the energy costs of external memory access as needed for table lookup of fully reconfigurable, sparse synaptic connectivity. Greater densities and energy efficiencies can be obtained by integrating them to replace the core of the external DRAM memory lookup in HiAER-IFAT flexible cognitive learning and inference systems with nanoscale memristor synapse arrays vertically interfacing with neuron arrays [32], and further optimize the vertically integrated circuits towards sub-pJ/spike overall energy efficiency in neocortical neural and synaptic computation and communication [33].

On-line unsupervised learning with event-driven contrastive divergence [34] using a wake-sleep modulated form of biologically inspired spike-timing dependent plasticity [35] produces generative models of spike-based neural representations offering a means to perform Bayesian inference in deep networks of large-scale spiking neuromorphic systems. The algorithmic advances in hierarchical deep learning and Bayesian inference harness the inherent stochastic nature of the computational primitives at the device level, such as the superior generalization and efficiency of learning of drop-connect emulating the pervasive stochastic nature of biological neurons and synapses [36], [37] by the Spiking Synaptic Sampling Machine (S3M) [38], [39].

## 3    DeepSouth

DeepSouth, the cortex emulator was designed for simulating large and structurally connected spiking neural networks in the lab of André van Schaik at the MARCS Institute, Western Sydney University, Australia. Inspired by observations from neurobiology, the fundamental computing unit is called a minicolumn, which consists of 100 neurons. Simulating large-scale, fully connected networks needs prohibitively large memory to store LUTs for point-to-point connections. Instead, we use a novel architecture, based on the structural connectivity in the neocortex, such that all the required parameters and connections can be stored in on-chip memory. The cortex emulator can be easily reconfigured for simulating different neural networks without any change in hardware structure by programming the memory. A hierarchical communication scheme allows one neuron to have a fan-out of up to 200k neurons. As a proof-of-concept, an implementation on a Terasic DE5 development kit was able to simulate 20 million to 2.6 billion leaky-integrate-and-fire (LIF) neurons in real time. When running at five times slower than real time, it is capable of simulating 100 million to 12.8 billion LIF neurons, which is the maximum network size on the chosen FPGA board, due to memory limitations. Larger networks could be implemented on larger FPGA boards with more external memory.

### 3.1    Strategy

### 3.1.1 Modular structure

The cortex is a structure composed of a large number of repeated units, neurons and synapses, each with several sub-types, as shown in Figure 5. A minicolumn is a vertical volume of cortex with about 100 neurons that stretches through all layers of the cortex [40]. Each minicolumn contains excitatory neurons, mainly pyramidal and stellate cells, inhibitory interneurons, and a large number of internal and external connections. The minicolumn is often considered to be both a functional and anatomical unit of the cortex [41], and we use this minicolumn with 100 neurons as the basic building block of the cortex emulator. The minicolumn in the cortex emulator is designed to have up to eight different programmable types of neurons. Note, the neuron types do not necessarily correspond to the cortical layers, but can be configured as such.

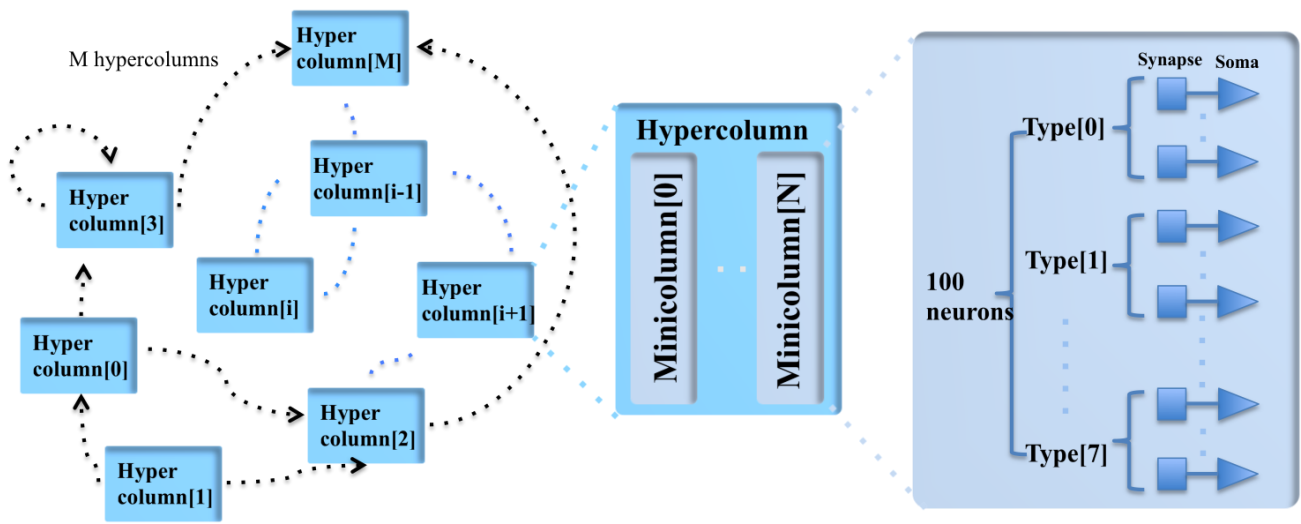

Figure 5. The modular structure of the cortex emulator. The basic building block of the cortex emulator is the minicolumn, which consists of up to eight different types of heterogeneous neurons (100 in total). The functional building block is the hypercolumn, which can have up to 128 minicolumns. The connections are hierarchical: hypercolumn-level connections, minicolumn-level connections and neuron-level connections.

In the mammalian cortex, minicolumns are grouped into modules called hypercolumns [42]. These are the building blocks for complex models of various areas of the cortex [43]. We therefore use the hypercolumn as a functional grouping for our emulator. The hypercolumn in our cortex emulator is designed to have up to 128 minicolumns. Similar to the minicolumns, the parameters of the hypercolumn are designed to be fully configurable.

### 3.1.2 Emulating dynamically

To solve the extensive computational requirement for simulating large networks, we use two approaches. Firstly, it is not necessary to implement all neurons physically on silicon and we can use time multiplexing to leverage the high-speed of the FPGA [44][45][46][47]. We can time-multiplex a single physical minicolumn (100 physical neurons in parallel) to simulate 200k time-multiplexed (TM) minicolumns, each one updated every millisecond. Limited by the hardware resources (mainly the memory), our cortex emulator was designed to be capable of simulating up to 200k TM minicolumns in real time and 1M ($2^{20}$) TM minicolumns at five times slower than real time, i.e., an update every 5 ms.

### 3.1.3 Hierarchical communication

Our cortex emulator uses a hierarchical communication scheme such that the communication cost between the neurons can be reduced by orders of magnitude. Anatomical studies of the cortex presented in [48] showed that cortical neurons are not randomly wired together and that the connections are quite structural. We chose to store the connection types of the neurons, the minicolumns, and the hypercolumns in a hierarchical fashion instead of individual point-to-point connections. In this scheme, the addresses of the events consist of hypercolumn addresses and minicolumn addresses. Both of them are generated on the fly with connection parameters according to their connection levels respectively. This method only requires several kilobytes of memory and can be easily implemented with on-chip SRAMs.

Inspired by observations from neurobiology, the communication between the neurons uses events (spike counts) instead of individual spikes. This arrangement models a cluster of synapses formed by an axon onto the dendritic branches of nearby neurons. The neurons of one type within a minicolumn all receive the same events, which are the numbers of the spikes generated by one type of neuron in the source minicolumns within a time step. One minicolumn has up to eight types of neurons, and each type can be connected to any type of neuron in the destination minicolumns. We impose that a source minicolumn has the same number of connections to all of the other minicolumns within the same hypercolumn, but that these can have different synaptic weights. The primary advantage of using this scheme is that it overcomes a key communication bottleneck that limits scalability for large-scale spiking neural network simulations.

Our system allows the events generated by one minicolumn to be propagated to up to 16 hypercolumns, each of which has up to 128 minicolumns, i.e., to 16×128×100 = 200k neurons. Each of these 16 connections has a configurable fixed axonal delay (from 1 ms to 16 ms, with a 1 ms step).

## 3.2 Hardware implementation

The cortex emulator was deliberately designed to be scalable and flexible, such that the same architecture could be implemented either on a standalone FPGA board or on multiple parallel FPGA boards.

As a proof-of-concept, we have implemented this architecture on a Terasic DE5 kit (with one Altera Stratix V FPGA, two DDR3 memories, and four QDRII memories) as a standalone system. Figure 6 shows its architecture, consisting of a neural engine, a Master, off-chip memories, and a serial interface.

The neural engine forms the main body of the system. It contains three functional modules: a minicolumn array, a synapse array, and an axon array. The minicolumn array implements TM minicolumns. The axon array will propagate the events generated by the minicolumns with axonal delays to the synapse array. In the synapse array, these events will be modulated with synaptic weights and will be assigned their destination minicolumn address. The synapse array will send these events to the destination minicolumn array in an event-driven fashion. Besides these functional modules, there is a parameter LUT, which stores the neuron parameters, connection types, and connection parameters. We will present the details of these modules in the following sections.

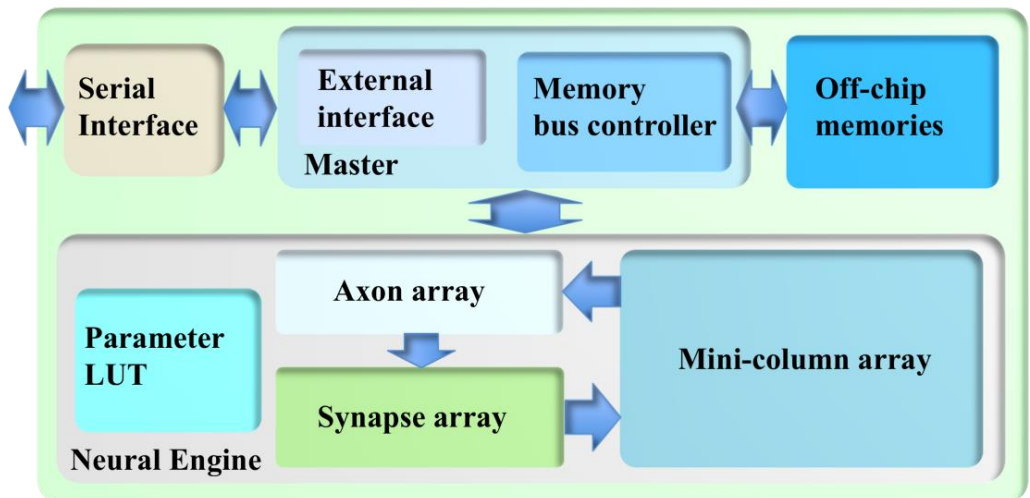

Figure 6. The architecture of the cortex emulator. The system consists of a neural engine, a Master, off-chip memories and a serial interface. The neural engine realizes the function of biological neural systems by emulating their structures. The Master controls the communication between the neural engine and the off-chip memories, which store the neural states and the events. The serial interface is used to interact with the other FPGAs and the host controller, e.g., PCs.

Because of the complexity of the system and large number of the events, each module in the neural engine was designed to be a slave module, such that a single Master has full control of the emulation progress. The Master has a memory bus controller that will control the access of the external memories. As we use time multiplexing to implement the minicolumn array, we will have to store the neural state variables of each TM neuron (such as their membrane potentials). These are too big to be stored in on-chip memory and have to be stored in off-chip memory, such as the DDR memory. Using off-chip memory needs flow control for the memory interface, which makes the architecture of the system significantly more complex, especially if there are multiple off-chip memories. The axon array also needs to access the off-chip memories for storing events.

The Master also has an external interface module that will perform flow control for external input and output. This module also takes care of instruction decoding. The serial interface is a high-speed interface, such as the PCIe interface, that communicates with the other FPGAs and the host PC. It is board-dependent, and in the work presented here, we use Altera's 10G base Phy IP.

### 3.2.1 Minicolumn array

The minicolumn array (Figure 7) consists of a neuron-type manager, an event generator, and the TM minicolumns, which have 100 parallel physical neurons. These neurons will generate positive (excitatory) and negative (inhibitory) post-synaptic currents (EPSCs and IPSCs) from input events weighted in the synapse array. These PSCs are integrated in the cell body (the soma). The soma performs a leaky integration of the PSCs to calculate the membrane potential and generates an output spike (post-synaptic spike) when the membrane potential passes a threshold, after which the membrane potential is reset and enters a refractory period. Events, i.e., spike counts, are sent to the axon array together with the addresses of the originating minicolumns, the number of connections, and axonal delay values for each connection (between two minicolumns).

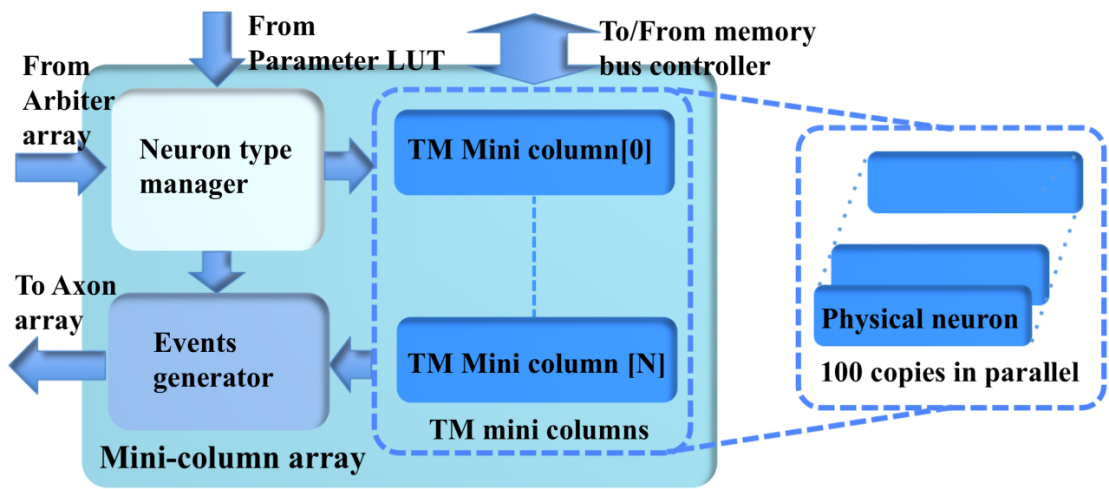

Figure 7. The structure of the minicolumn array.

### 3.2.2 Axon array

The axon array propagates the events from the minicolumn array or from the external interface to the synapse array, using programmable axonal delays. To implement this function on hardware, we used a two-phase scheme comprising a TX-phase and an RX-phase. In the TX-phase, the events are written into different regions of the DDR memories according to their programmable axonal delay values. In the RX-phase, for each desired axonal delay value, the events are read out from the corresponding region of the DDR memories stochastically, such that their expected delay values are approximately equal to the desired ones.

### 3.2.3 Synapse array

The synapse array emulates the function of biological synaptic connections: it modulates the incoming events from the axon array with synaptic weights and generates destination minicolumn addresses for them. These events are then sent to the TM minicolumns. The synapse array only performs the linear accumulation of synaptic weights of the incoming events, whereas the exponential decay is emulated by the PSC generator in the neuron.

### 3.2.4 Master

The Master plays a vital role in the cortex emulator: it has complete control over all the modules in the neural engine such that it can manage the progress of the simulation. This mechanism effectively guarantees no event loss or deadlock during the simulation. The Master slows down the simulation by pausing the modules that are running quicker than other modules. The Master has two components (Figure 6): a memory bus controller and an external interface. The memory bus controller has two functions: interfacing the off-chip memory with Altera IPs, and managing the memory bus sharing between the minicolumn array and the axon array.

The external interface module will control the flow of the input and output of events and parameters. The main input to this system are events, which are sent to the minicolumn array via the axon array. This module also performs instruction decoding such that we can configure the parameter LUT and the system registers. The outputs of the emulator are individual spikes (100 bits, one per neuron) and events generated by the minicolumns.

### 3.3 Programming API

Along with the hardware platform, we also developed a simple application programming interface (API) in Python that is similar to the PyNN programming interface [49]. This API is very similar to the high-level object-oriented interface that has been defined in the PyNN specification: it allows users to specify the parameters of neurons and connections, as well as the network structure using Python. This will enable the rapid modelling of different topologies and configurations using the cortex emulator. This API allows monitoring of the generated spikes in different hypercolumns. As future work, we plan to provide full support for PyNN scripts and incorporate interactive visualization features on the cortex emulator.

### 4 BrainScaleS

The BrainScaleS neuromorphic system has been developed at the University of Heidelberg in collaboration with the Technical University Dresden and the Frauenhofer IZM in Berlin. The BrainScaleS neuromorphic system is based on the direct emulation of model equations describing the temporal evolution of neuron and synapse variables. The electronic neuron and synapse circuits act as physical models for these equations. Their measurable electrical quantities represent the variables of the model equations, thereby implicitly solving the related differential equations.

The current first generation BrainScaleS system implements the Adaptive Exponential Integrate-and-Fire Model [50]. All parameters are linearly scaled to match the operating conditions of the electronic circuits. The membrane voltage range between hyperpolarization, i.e., the reset voltage in the model, and depolarization (firing threshold) is approximately 500 mV. Time, being a model variable as well, can also be scaled in a physical model. In BrainScaleS, this is used to accelerate the model in comparison to biological wall time. It uses a target acceleration factor of $10^4$. This acceleration factor has a strong influence on most of the design decisions, since the communication rate between the neurons scales directly with the acceleration factor, i.e., all firing rates are also $10^4$ higher than those in biological systems. The rationale behind the whole communication scheme within the BrainScaleS system is based on these high effective firing rates.

In the BrainScaleS system the whole neuron, including all its synapses, is implemented as a continuous-time analog circuit. Therefore, it consumes a substantial silicon area. To be able to implement networks larger than a few hundred neurons, a multichip implementation is necessary.

Due to the high acceleration factor, this requires very high communication bandwidth between the individual Application Specific Integrated Circuits (ASICs) of such a multichip system. BrainScaleS uses wafer-scale integration to solve this problem. Figure 8 shows a photograph of a BrainScaleS wafer module integrating an uncut silicon wafer with 384 neuromorphic chips. Its neuromorphic components will be described in the remainder of this section, which is organized as follow: subsection 4.1 explains the basic neural network circuits, while 4.2 details the wafer-scale communication infrastructure. Finally, the wafer-scale integration is described in subsection 4.3.

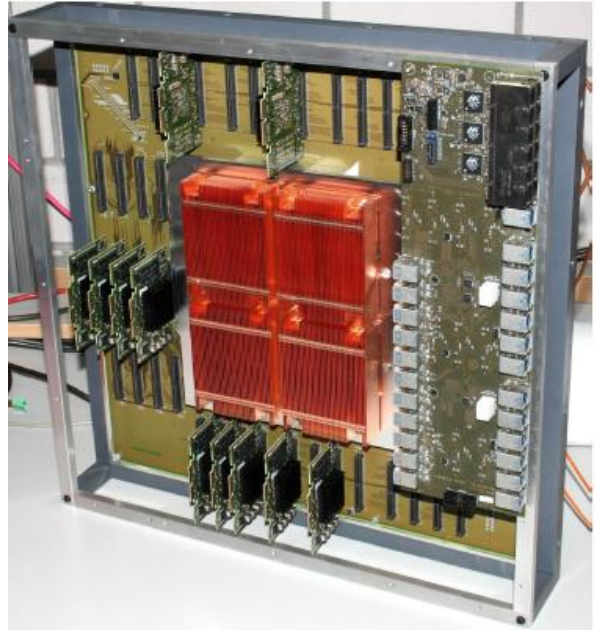

Figure 8. Photograph of a BrainScaleS wafer module with part of the communication boards removed to show the main PCB. The wafer is located beneath the copper heat sink visible in the center.

## 4.1 HICANN ASIC

At the center of the BrainScaleS neuromorphic hardware system is the High-Input Count Analog Neuronal Network Chip (HICANN) ASIC. Figure 9 shows a micro-photograph of a single HICANN die. The center section contains the symmetrical analog network core: two arrays with synapse circuits enclose the neuron blocks in the center. Each neuron block has an associated analog parameter storage. The communication network which surrounds the analog network core will be described in Section 4.2. The first HICANN prototype is described in [51]. The current BrainScaleS system is built upon the third version of the HICANN chip, which is mostly a bug-fix version. A second generation BrainScaleS system based on a smaller manufacturing process feature size is currently under development. It shrinks the design geometries from 180 nm to 65 nm and improves part of the neuron circuit [52], adds hybrid plasticity [53] and integrates a high-speed Analog-to-Digital Converter (ADC) for membrane voltage readout. This paper refers to the latest version of the first generation HICANN chip as it is currently used in the BrainScaleS system.

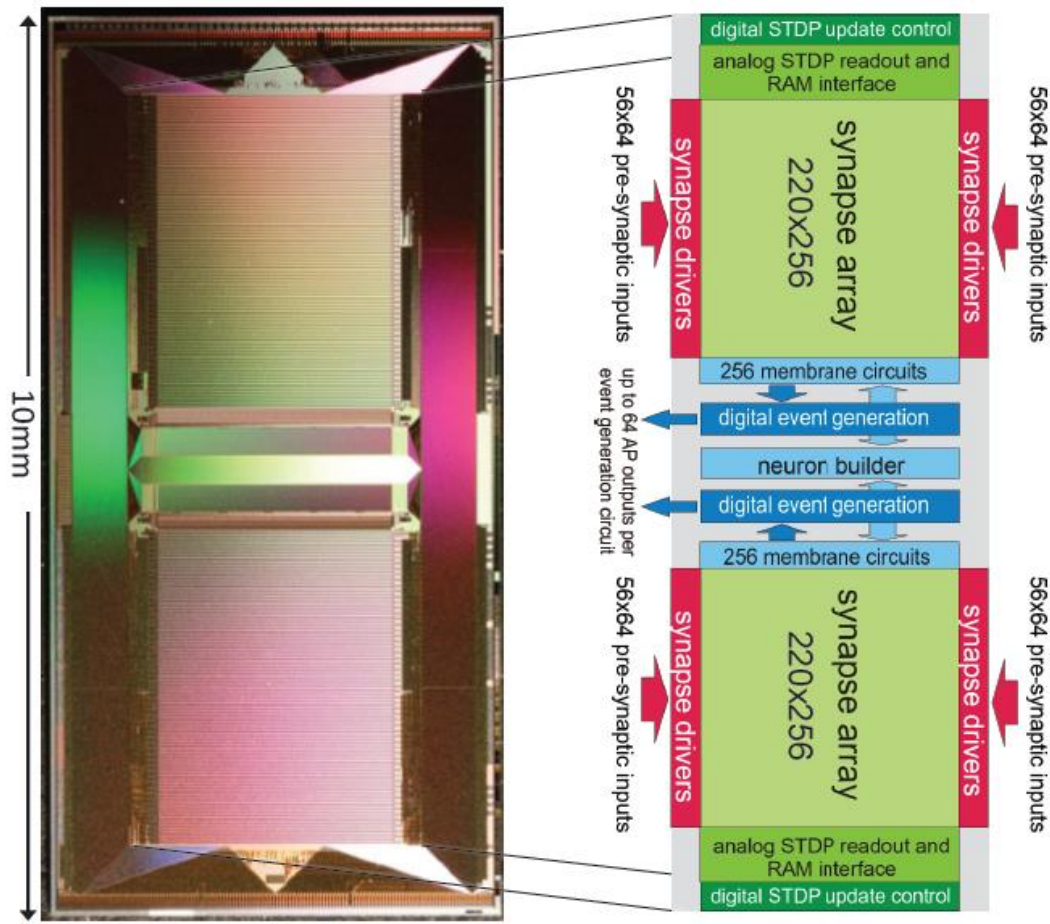

Figure 9. Photograph of a single HICANN die. The enlargement shows a block diagram of the analog network core located in the center of the die.

Figure 10 shows its main components. The analog network core contains the actual analog neuron and synapse circuits. The network core communicates by generating and receiving digital event signals, which correspond to biological action potentials. One major goal of HICANN is a fan-in per neuron of more than 10k pre-synaptic neurons. To limit the input ports of the analog core to a manageable number, time-multiplexing is used for the event communication: each communication channel is shared by 64 neurons. Each neuronal event transmits therefore a 6-bit number while time is coded by itself, i.e., the event communication happens in real-time related to the emulated network model. This communication model is subsequently called Layer 1 (L1).

The Layer 2 (L2) encoding is used to communicate neural events in-between the wafer and the host compute cluster. Due to the latencies involved with long-range communication, it is not feasible to use a real-time communication scheme. A latency of 100ns would translate to a 1ms delay in biological wall time using an acceleration factor of $10^4$. Therefore, a protocol based on packet-switching and embedded digitized time information is used for the L2 host communication.

The grey areas labeled 'digital core logic' and 'STDP top/bottom' are based on synthesized standard-cell logic. In Figure 9 they are not visible, because the L1 communication lines are located on top of the standard cell areas (Section 4.2). They occupy the whole area surrounding the analog network core. The core itself uses a full-custom mixed-signal implementation, as do the repeaters,

Serializer/De-Serializer (SERDES) and Phase-Locked Loop (PLL) circuits. The only purely analog components are two output amplifiers. They allow direct monitoring of two selectable analog signals.

The RX- and TX circuits implement a full-duplex high-speed serial link to communicate with the wafer module. The thick red arrows represent the physical L1 lanes. Together with the vertical repeater circuits and the synapse and vertical-horizontal crossbars, they form the L1 network (Section 4.2).

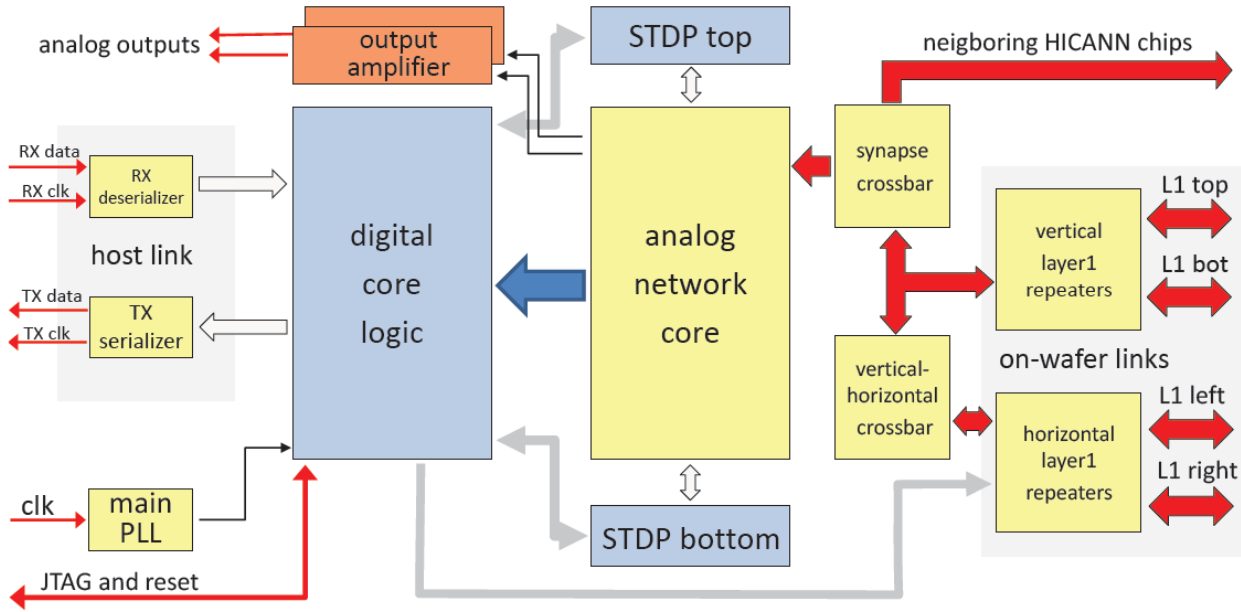

Figure 10. Block diagram of the HICANN chip.

### 4.1.1 Neuron Circuits

The HICANN neuron circuits are based on the AdEx model [54]. Details of the circuit implementation of this model and measurement results of the silicon neuron can be found in [55] and [56]. Figure 11 shows the basic elements of a membrane circuit. A row of 256 membrane circuits is located adjacent to each of the two synapse arrays. Within the 2-dimensional synapse arrays, each membrane circuit has one column of 220 synapses associated with it. To allow for neurons with more than 220 inputs, up to 64 membrane circuits can be combined to one effective neuron.

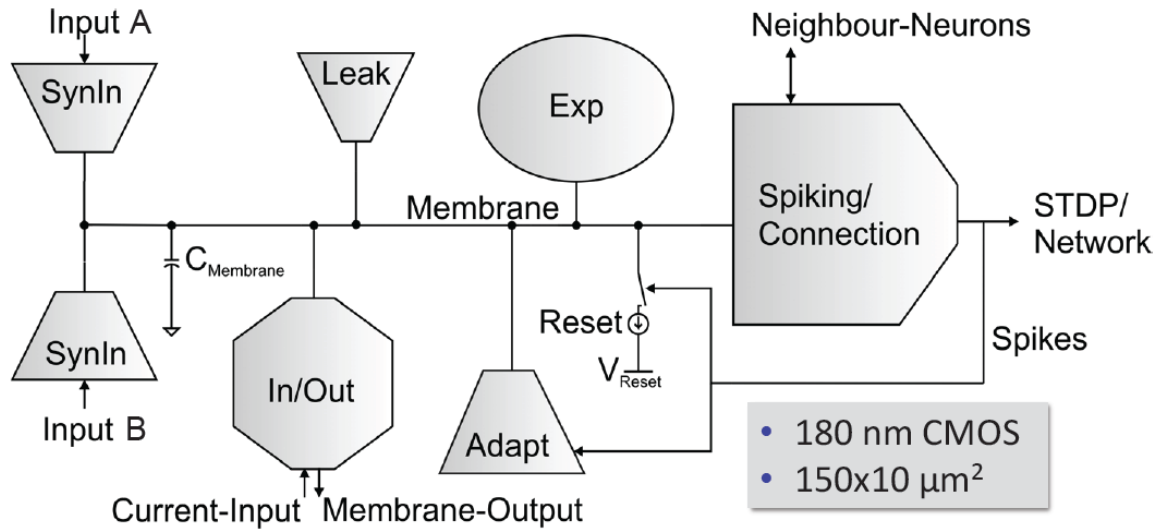

Figure 11. Block diagram of a HICANN membrane circuit. The model neurons are formed by interconnecting groups of membrane circuits.

The circuit named 'neuron builder' in Figure 9 is responsible for interconnecting the membrane capacitances of the membrane circuits that will operate together as one model neuron. Each membrane circuit contains a block called 'Spiking/Connection', which is able to generate a spike if the membrane crosses its threshold voltage. This spike generation circuit can be individually enabled for each membrane circuit. In neurons built by interconnecting a multitude of membrane circuits, only one spike generation circuit is enabled. The output of the spike generation circuit is a digital pulse lasting for a few nanoseconds. It is fed into the digital event generation circuit located below each neuron row as shown in Figure 9.

Additionally, each membrane circuits sends the spike signal back into its synapse array column, where it is used as the post-synaptic signal in the temporal correlation measurement between pre- and post-synaptic events. Within a group of connected membrane circuits, their spike signals are connected as well. Thereby, the spike signal from the single enabled spike generation circuit is reflected in all connected membrane circuits and driven as post-synaptic signal in all synaptic columns belonging to the connected membrane circuits. The neuron uses 23 analog parameters for calibration. They are split in 12 voltage parameters, like the reversal potentials or the threshold voltage of the neuron, and 11 bias currents. These parameters are stored adjacent to the neurons in an array of analog memory cells. The memory cells are implemented using single-poly floating-gate technology.

Each membrane circuit is connected to 220 synapses by means of two individual synaptic input circuits. Each row of synapses can be configured to use either of them, but not both simultaneously. Usually they are configured to model the excitatory and inhibitory inputs of the neuron.

The synaptic input uses a current-mode implementation. An operational amplifier (OP) acts as an integrator located at the input (see Figure 12). It keeps the voltage level of the input constant at $V_{syn}$. Each time a synapse receives a pre-synaptic signal, it sinks a certain amount of current for a fixed time interval of nominal 4ns. To restore the input voltage level to $V_{syn}$, the integrator has to source the corresponding amount of charge through its feedback capacitor.

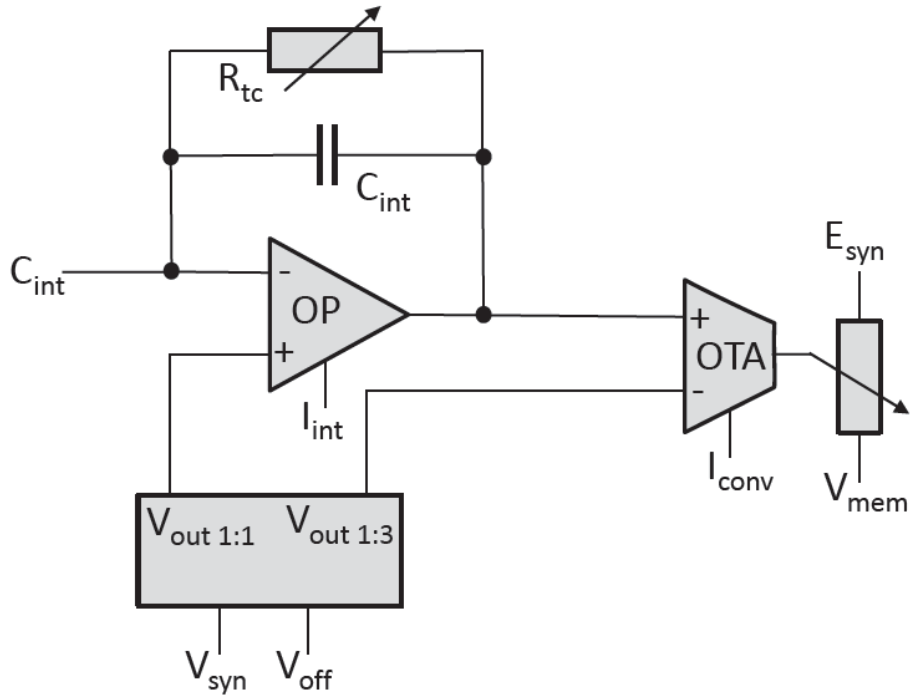

Figure 12. Simplified circuit diagram of the synaptic input of the neuron.

The second amplifier is an Operational Transconductance Amplifier (OTA) that converts the voltage over the feedback capacitor $C_{int}$ into a current that is subsequently used to control a current controlled resistor connecting the reversal potential to the membrane. The exponential decay of the synaptic conductance is generated by the adjustable resistor $R_{tc}$ in parallel to $C_{int}$. The rise-time of the synaptic conductance is controlled by the bias current $I_{int}$ of the integrator. The bias current of the OTA, $I_{conv}$, sets the ratio between the conductance and the synaptic current.

### 4.1.2 Synapse array and drivers

Figure 13 shows the different components of the synapse. The control signals from the synapse drivers run horizontally through the synapse array, orthogonal to the neuron dendritic current inputs (synaptic input). The input select signal statically determines which current input of the membrane circuits the synapses use. It can be set individually for each row of synapses.

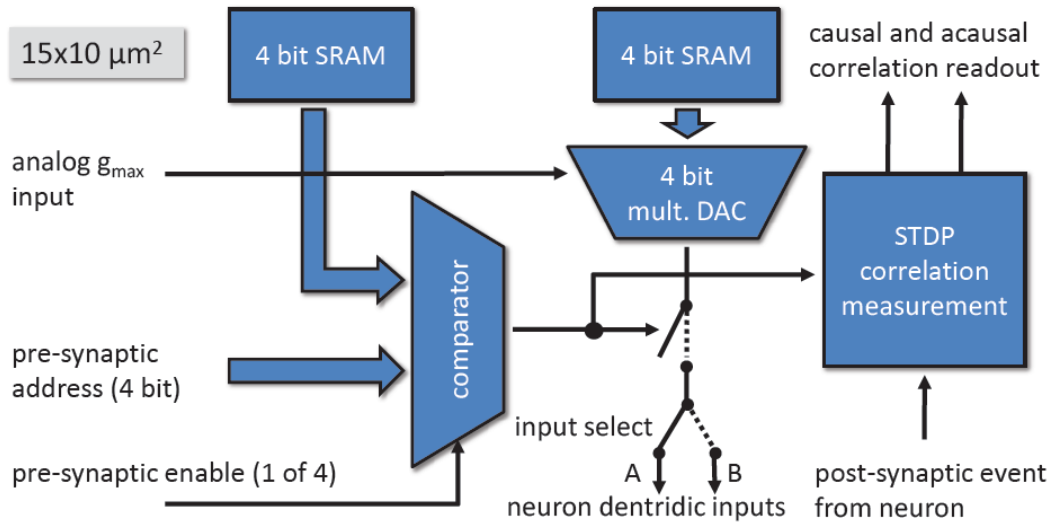

Figure 13. Block diagram of the synapse.

A synapse is selected when the 4-bit pre-synaptic address matches the 4-bit address stored in the address memory while the pre-synaptic enable signal is active. The synaptic weight of the synapse is realized by a 4-bit Digital-to-Analog Converter (DAC) and a 4-bit weight memory. The output current of the DAC can be scaled with the row-wise bias $g_{max}$. During the active period of the pre-synaptic enable signal, the synapse sinks the current set by the weight and $g_{max}$.

By modulating the length of the pre-synaptic enable signal, the total charge sunk by the synapse can be adjusted for each pre-synaptic spike. The synapse drivers use this to implement short-time plasticity.

Each synapse contains in addition to the current sink, a correlation measurement circuit to implement Spike Timing Dependent Plasticity (STDP) [57]. The pre-synaptic signals are generated within the synapse drivers, located to the left and right side of the synapse array in Figure 9. Each synapse drivers controls two adjacent rows of synapses.

## 4.2   Communication infrastructure

The analog network core described in the previous section implements a total of 220 synapse driver circuits. Each synapse driver receives one L1 signal.

The physical L1 connections use the topmost metal layer (metal6) for vertical lines and the metal layer beneath for horizontal lines. Since the synapse array uses all metal layers, the only physical space for L1 connections in HICANN is the area surrounding the analog network core and the center of the analog network core, above the digital event generation and neuron builder circuits shown in Figure 10.

There are 64 horizontal lines in the center and 128 to the left and to the right of the analog network core. Due to the differential signaling scheme used, each line needs two wires; the total number of wires used is 640, each wire pair using a maximum signal bandwidth of 2 Gbits$^{-1}$. This adds up to a total bandwidth of 640 Gbits$^{-1}$.

#### 4.2.1 Wafer-scale integration

To increase the size of an emulated network beyond the number of neurons and synapses of a single HICANN chip, some kind of chip-to-chip interconnect is necessary. Any 3D integration technique [58] will allow to stack only a few chips reliably. Therefore, on its own, 3D stacking is not the solution to scale HICANN to large network sizes.

Packaging the HICANN chips for flip-chip PCB mounting and soldering them to carrier boards would be a well-established option. Driving the additional capacity and inductance of two packages and a PCB trace would make the simple asynchronous differential transmission scheme unfeasible. Therefore, the area and power used by the communication circuits would most likely increase. To solve the interconnection and packaging problems a different solution was chosen: wafer-scale integration. Wafer-scale integration, i.e., the usage of a whole production wafer instead of dicing it into individual reticles, is usually not supported by the semiconductor manufacturing processes available for university projects. Everything is optimized for mass-market production, where chip sizes rarely reach the reticle limit. Stitching individual reticles together on the top metal layer was therefore not available for HICANN. In the 180 nm manufacturing process used, eight HICANN dies fit on one reticle. So two problems had to be solved: how to interconnect the individual reticles on the wafer and how to connect the whole wafer to the system.

Within a reticle, the connections between the L1 lines of neighboring HICANN chips are made directly by top layer metal. With stitching available, interconnects on top layer metal would allow to continue the L1 lines across reticle boundaries. But, the problem of the wafer to PCB connection would still remain. The solution employed in BrainScaleS solves both connection problems and can be used with any silicon wafer manufactured in a standard CMOS process. It is based on the post-processing of the manufactured wafers. A multi-layer wafer-scale metalization scheme has been developed by the Frauenhofer IZM in Berlin. It uses a wafer-scale maskset with $\mu$m resolution.

After some initial research, a minimum pad window size of $5\times5\mu m^2$ was chosen. $15\times15\mu m^2$ copper areas have proven to connect reliably to these pad windows. In Figure 14, a photograph with all post-processing layers applied to a wafer of HICANN chips is shown. The cut-out with the maximum enlargement in the top left corner shows the dense interconnects linking the L1 lines of two HICANN chips in adjacent reticles. They use a pitch of 8.4$\mu$m[1]. Due to the larger size of the post-processing to wafer contacts, some staggering is necessary.

A second, much thicker post-processing layer is used to create the large pads visible in the eight columns in the inner area of the reticle. They solve the second problem related to wafer-scale integration: the wafer to PCB connection. Since the L1 connections are only needed at the edge of the reticle, the whole inner area is free to redistribute the HICANN IO and power lines and connect them to the regular contact stripes visible in the photograph.

---

[1] The post-processing process has proven to be reliable down to a pitch of 6 $\mu$m. Since this density was not necessary, the pitch was relaxed a bit.

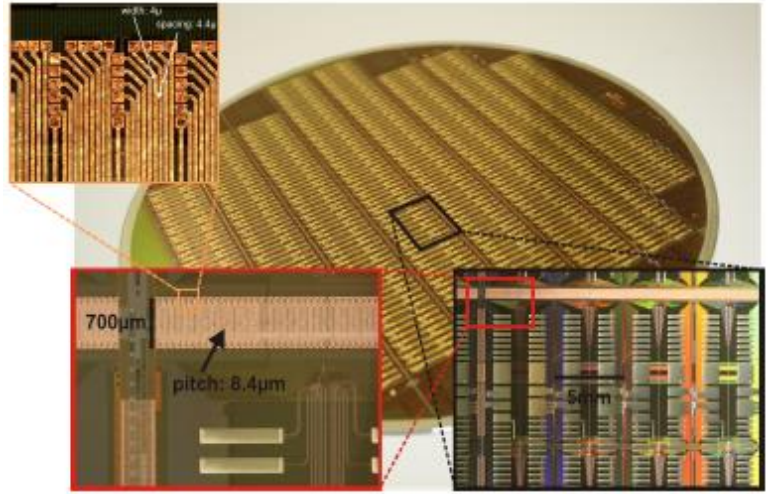

Figure 14. L1 post processing. Main picture: Photograph of a post-processed wafer. Enlargements: bottom right: top-left corner of a reticle. The large stripes connect the wafer to the main PCB. Bottom left: the L1 connections crossing the reticle border. Top left: 4 μm wide individual metal traces of the L1 connections terminating at the pad-windows of the top reticle.

## 4.3 Wafer module

A photograph of a partial assembled wafer module is shown in Figure 9. The visible components are the main PCB with the wafer in the center, beneath the copper heat-sink. There are 48 connectors for the Communication Subgroup (CS) surrounding the wafer. Each CS provides the Low-Voltage Differential Signaling (LVDS) and Joint Test Action Group (JTAG) signals for one reticle[2]. Each wafer module uses 48 1Gbit/s Ethernet links to communicate with the host compute cluster via an industrial standard Ethernet switch hierarchy.

The links are provided by four IO boards mounted on top of the CSs. In Figure 9, only one of them is in place to avoid blocking the view on the CSs. Twelve RJ45 Ethernet Connectors (RJ45s) can be seen in the upper right corner of the IO board. The remaining connectors visible are for future direct wafer-to-wafer networking.

### 4.3.1 Wafer-PCB connection

As shown in Figure 14, the HICANN wafer is connected to the main PCB by a contact array formed by the wafer post-processing. The stripes visible in the figure have a width of 1.2 mm and a pitch of 400 $\mu m$. A mirror image of theses stripes is placed on the main PCB.

The connection between both stripe patterns is then formed by Elastomeric Connectors[3] with a width of 1 mm and a density of 5 conducting stripes per mm. Therefore, they do not have to be precisely aligned to the PCB or the wafer, only the wafer and the PCB have to be aligned to each other with about 50 $\mu m$ accuracy. This is achieved by placing the wafer in an aluminum bracket, which is fastened to an aluminum frame located on the opposite side of the main PCB by eight precision screws.

---

[2] The CS was developed by Technical University Dresden

[3] The connectors used are from Fujipoly, Datasheet Nr. FPDS 01-34 V6

To place the Elastomic Connectors at the correct positions, they are held by a thin, slotted FR4 mask which is screwed to the main PCB. A special alignment tool has been developed to optically inspect and correct the wafer to PCB alignment during the fastening of the screws. The optimum pressure is achieved by monitoring the electrical resistance of a selected set of wafer to PCB contacts during the alignment and fastening process. After the wafer is correctly in place, the bracket is filled with nitrogen and sealed.

## 4.4   Summary of the BrainScaleS system

The BrainScaleS system, based upon the BrainScaleS wafer module described in the previous sections, is an attempt to simultaneously fulfill the following requirements:

- continuous time analog implementation of neurons and synapses
- programmable analog parameters to calibrate neurons according to models from computational neuroscience
- programmable topology to implement said target models
- fully parallel, per synapse long-term plasticity
- scalability across single chip limits using wafer-scale integration
- accelerated operation

By compressing the model timescale by several orders of magnitude it allows to model processes like learning and development in seconds instead of hours. It will make parameter searches and statistical analysis possible in all kind of models. Some results of accelerated analog neuromorphic hardware are reported in [59][60]–[62].

## 5   <u>Dynap-SEL</u>: A multi-core spiking chip for models of cortical computation

Novel mixed-signal multi-core neuromorphic processors that combine the advantages of analog computation and digital asynchronous communication and routing have been recently designed and fabricated in both standard 0.18 $\mu$m CMOS processes [63] and advanced 28 nm Fully-Depleted Silicon on Insulator (FDSOI) processes [64] in the lab of Giacomo Indiveri at University of Zurich, Switzerland. The analog circuits used to implement neural processing functions have been presented and characterized in [65]. Here we describe the routing and communication architecture of the 28 nm "Dynamic Neuromorphic Asynchronous Processor with Scalable and Learning" (Dynap-SEL) device, highlighting both its run-time network re-configurability properties and its on-line learning features.

### 5.1   The Dynap-SEL neuromorphic processor

The Dynap-SEL chip is a mixed-signal multi-core neuromorphic processor that comprises four neural processing cores, each with 16×16 analog neurons and 64 4-bit programmable synapses per neuron, and a fifth core with 1×64 analog neuron circuits, 64×128 plastic synapses with on chip learning circuits, and 64×64 programmable synapses. All synaptic inputs in all cores are triggered by incoming Address Events (AEs), which are routed among cores and across chips by asynchronous Address-Event Representation (AER) digital router circuits. Neurons integrate synaptic input currents and eventually produce output spikes, which are translated into AEs and routed to the desired destination via the AER routing circuits.

### 5.1.1 Dynap-SEL routing architecture

The Dynap-SEL routing architecture is shown in Figure 15. It is composed of a hierarchy of routers at three different levels that use both source-address and destination-address routing. The memory structures distributed within the architecture to support the heterogeneous routing methods employ both Static Random Access Memory (SRAM) and Ternary Content Addressable Memory (TCAM) memory elements (see [63] for a detailed description of the routing schemes adopted). To minimize memory requirements, latency, and bandwidth usage, the routers follow a mixed-mode approach that combines the advantages of mesh routing (low bandwidth usage, but high latency), with those of hierarchical routing (low latency, but high bandwidth usage) [66]. The asynchronous digital circuits in Dynap-SEL route spikes among neurons both within a core, across cores, and across chip boundaries. Output events generated by the neurons can be routed to the same core, via a Level-1 router R1; to other cores on the same chip, via a Level-2 router R2; or to cores on different chips, via a Level-3 router R3. The R1 routers use source-address routing to broadcast the address of the sender node to the whole core, and rely on the TCAM cells programmed with appropriate tags to accept and receive the AE being transmitted. The R2 routers use absolute destination-address routing in a 2D tree to target the desired destination core address. The R3 routers use relative destination address-routing to target a destination chip at position ($\Delta x$, $\Delta y$). The memory used by the routers to store post synaptic destination addresses is implemented using 8.5k 16-bit SRAM blocks distributed among the Level-1, -2, and -3 router circuits.

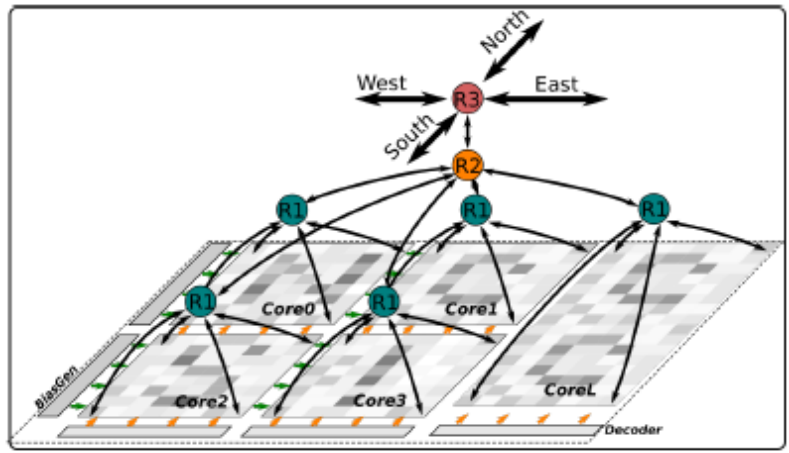

Figure 15. Dynap-SEL with hierarchical routers. Each Dynap-SEL comprises four TCAM-based non-plastic cores and one plastic core. The hierarchical routing scheme is implemented on three levels for intra-core (R1), inter-core (R2), and inter-chip (R3) communication.

In each non-plastic core, 256 analog neurons and 16k asynchronous TCAM-based synapses are distributed in a 2D array. Each neuron in this array comprises 64 synapses with programmable weights and source-address tags. Thanks to the features of the TCAM circuits, there are $2^{11}$ potential input sources per synapse. Synaptic input events are integrated over time by a Differential Pair Integrator (DPI) linear integrator circuit [67] that exhibits dynamics with biologically realistic time constants (e.g., of the order of tens of milliseconds). Thanks to its modularity, scalability, and on-chip programmable routers, the Dynap-SEL can be integrated in a chip array of up to 16×16 chips, allowing all-to-all connectivity among all neurons in the array. This enables the implementation of a wide range of connections schemes, without requiring any additional external mapping, memory, or computing support. By following the parallel AER protocol, it is possible to establish direct communications between Dynap-SEL chips and other AER sensors and computing devices, enabling the construction of large-scale sensory processing systems.

The TCAM and SRAM circuits are subdivided into small memory blocks and embedded within the neuron and synapse arrays. Given that memory and computation are co-localized, each memory access operation is extremely efficient in terms of power consumption, compared to the classical scheme of accessing large TCAM/SRAM blocks placed at longer distances. As there are only local and sparse memory access operations, the requirement of memory bandwidth is also much lower than in traditional von Neumann architecture. In addition to the power and bandwidth benefits, this distributed heterogeneous memory architecture lends itself well to the exploitation of emerging memory technologies, e.g., by replacing the CMOS Content Addressable Memory (CAM) or SRAM cells with nano-scale Resistive Random Access Memories (ReRAMs) or memristors [68],[69]. By careful design of the analog circuits, we were able to achieve an extremely compact layout. This allowed us to implement multiple physical analog neurons for true parallel computing, rather than using time-multiplexing to share the computing resources of digital neural processing blocks (e.g., as it is done in [8]). In the 28 nm process used, the analog neurons occupy around 5% of whole chip area.

### 5.1.2 The plastic core

This core in the Dynap-SEL device comprises 64 analog neurons. Each neuron has 128 mixed-signal plastic synapses, 64 mixed-signal non-plastic synapses, and 4 linear synapse circuits. The plastic and non-plastic synapses have synaptic weight parameters with 4-bit resolution, while the linear synapse circuits have four independent set of parameters that can be set by a 12-bit programmable bias-generators. These parameters can be used to change the synaptic weights, the time constants, or the type of excitatory/inhibitory synapse. Furthermore, each of the linear synapses can be time-multiplexed to represent many different synaptic inputs that have same weight and temporal dynamics (e.g., a 1 KHz input spike train could represent 1000 1 Hz synaptic inputs). The block diagram of the plastic core is shown in Figure 16.

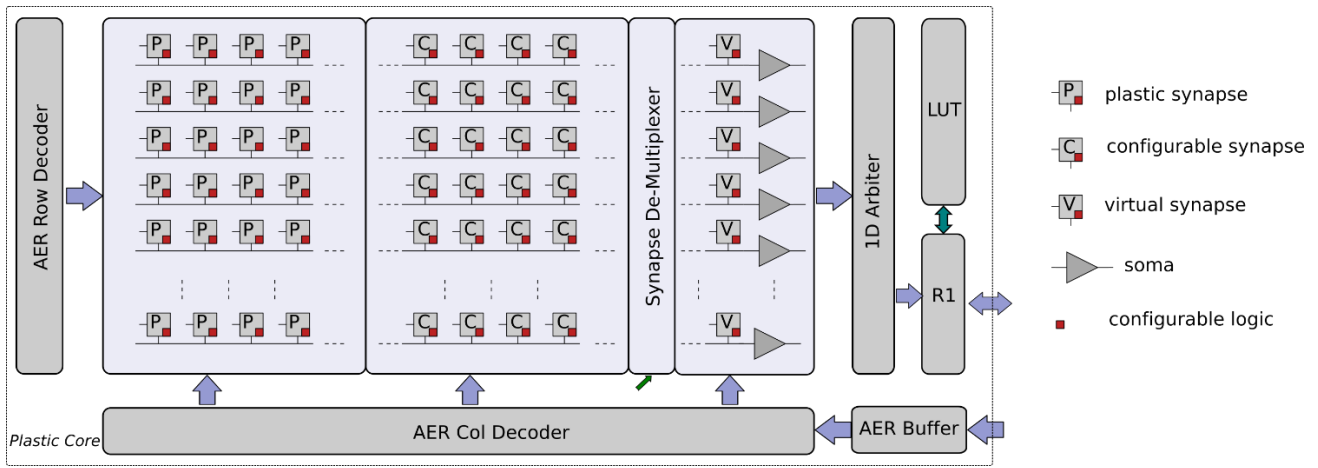

Figure 16. Plastic core architecture. It comprises a 64×128 array of plastic synapses, a 64×64 array of non-plastic synapses, a 64×4 array of time-multiplexed linear synapses, a synapse row de-multiplexer, and 64 neurons adaptive I&F neurons. Each synapse has digital memory and configuration logic circuits. In addition, there are AER input Row/Column decoders, a 1D AER output arbiter, a local router R1, and a corresponding LUT for routing.

The weight update mechanism of the plastic synapses is governed by a 4-bit up/down counter, which is used to implement the spike-driven learning rule proposed in [70] (see also [9] for a detailed description of the learning circuits involved). Local digital latches are placed in each synapse to set the synapse type (excitatory/inhibitory), to impose a user-specified weight value, to enable/disable local weight monitor circuits, to enable/disable the learning, or to enable/disable the access to the input broadcast line. Non-plastic synapse circuits are a simplified version of the plastic ones, which do not have the on-line learning mechanism, but share all other features. A local pulse-to-current converter is placed in each synapse to convert the fast AER input events into longer current pulses with tunable pulse width, to provide an additional degree of control over the synaptic weight amplitude range. The long weighted current pulses are then summed and integrated by DPI filter circuits [67] located at the side of the synapse array. The filtered output of the weighted sum of currents is then fed into their corresponding neuron circuit. A synapse demultiplexer allows the user to assign more synapse rows to targeted neurons, to increase the size of the input space/neuron (at the cost of decreasing the number of active neurons). It is possible to merge at most eight rows of synapses to achieve a fan-in of 1k plastic synapses and 512 non-plastic ones per neuron, for a network of eight usable neurons.

The spikes generated by the neurons get encoded into AEs by the 1D arbiter. The LUT is used to append a destination chip and core address to each AE. There can be up to eight different copies of an AE with eight different destination addresses, in order to increase the fan-out of each neuron and allow it to target up to 8 different chips and maximum 32 cores. The local R1 router then routes the AEs to the corresponding destinations. Once an AE reaches its destination core, the address gets broadcast to the whole core (which comprises 256 neurons). So, in principle it is possible to achieve a maximum fan-out of 8k destinations.

## 5.2    Conclusion of the Dynap-SEL system

Thanks to the flexibility of the memory-optimized routing system adopted, the system is highly scalable [71]. Resources from different chips can be easily combined and merged. Plastic cores from up to 4×4 chips can be merged together to build a larger core. For example, by merging plastic cores from 16 chips, a plastic core with 128×1k plastic synapses and 1k neurons, or a plastic core with 1k×128 plastic synapses and 128 neurons can be configured. The implementation of asynchronous memory control allows the on-line re-configuration of the routing tables. This in turn allows the implementation of structural plasticity or evolutionary algorithms.

By using multiple physical circuits to carry out parallel computation, and by implementing many small and distributed memory structures embedded within the mixed-signal neural processing fabric, this architecture eliminates, by design, the von Neumann bottleneck problem at the source. Memory and computation are co-localized at all levels of the hierarchy (Figure 17). In the current design most of the silicon real estate is occupied by the digital SRAM and TCAM memory cells. The use of emerging memory technologies, such as ReRAM, will allow to dramatically reduce the size of the design, and to integrate on the same area larger numbers of synapses and neurons.

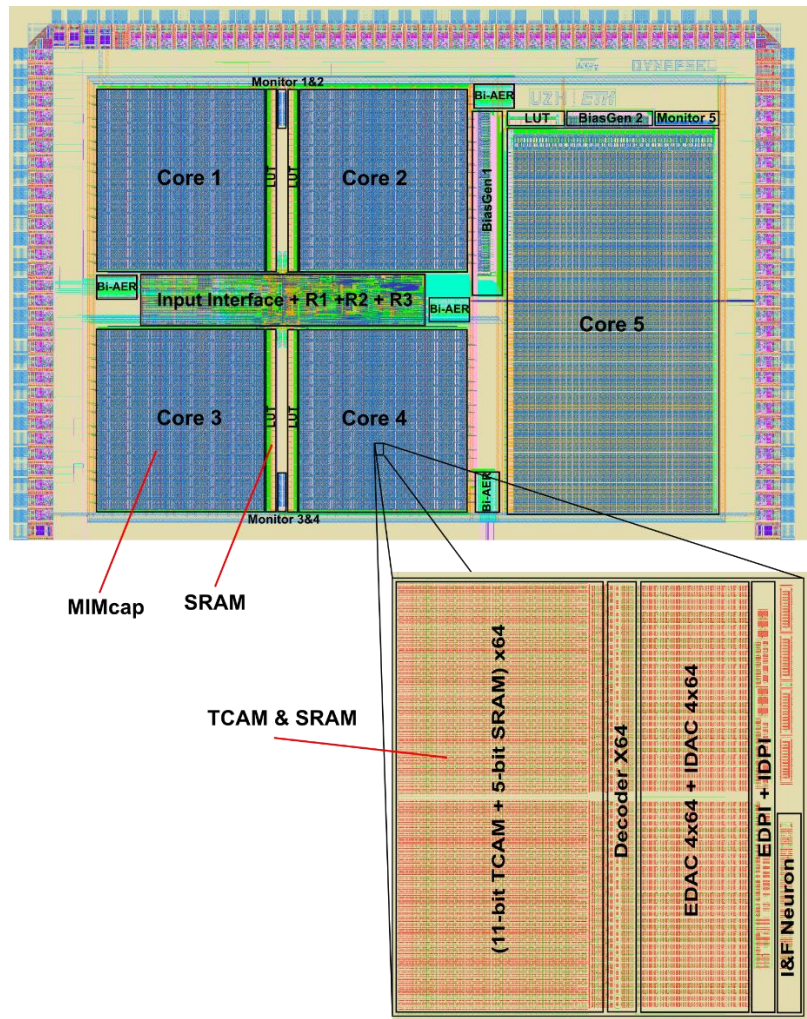

Figure 17. Dynap-SEL chip, fabricated using a 28 nm FDSOI process. It occupies an area of 7.28 mm$^2$ and comprises four non-plastic cores and one plastic core. Each non-plastic core has 256 analog I&F neurons and 64k TCAM-based synapses arranged in 2D array (each synapse address is identified by an 11-bit TCAM and each synapse type and weight is identified by a 5-bit SRAM). Each plastic core has 64 analog I&F neurons, 8k digital plastic synapses, 4k digital non-plastic synapses, and 256 virtual synapses.

## 6 The 2DIFWTA chip: a 2D array of integrate-and-fire neurons for implementing cooperative-competitive networks

A competitive network typically consists of a group of interacting neurons, which compete with each other in response to an input stimulus. The neurons with the highest response suppress all other neurons to win the competition. Competition is achieved through a recurrent pattern of connectivity involving both excitatory and inhibitory connections. Cooperation between neurons with similar response properties (e.g., close receptive field or stimulus preference) is mediated by excitatory connections. Cooperative-competitive networks perform not only common linear operations but also complex nonlinear operations. The linear operations include analog gain (linear amplification of the feedforward input, mediated by the recurrent excitation and/or common mode input), and locus invariance [72]. The nonlinear operations include non-linear selection or soft winner-take-all (WTA) behavior [73], [74], [75], signal restoration [74], [76], and multi-stability [73], [74], [75].

We were interested in exploring the computational properties of cooperative-competitive neural networks in both the mean rate and time domain. To this end, we designed a VSLI chip with a dedicated architecture for implementing a single 2D cooperative-competitive network or multiple 1D cooperative-competitive networks.

## 6.1 Chip description

The 2DIFWTA (2D Integrate-and-Fire Winner-Take-All) chip was implemented using a standard 0.35 $\mu m$ four-metal CMOS technology (Figure 18). It comprises a two dimensional array of 32×64 (2048) I&F neurons. Each neuron (Figure 19) receives inputs from AER synapses (two excitatory and one inhibitory) and local excitatory synapses. The local connections implement recurrent cooperation for either a two-dimensional or 32 mono-dimensional WTA networks. Cooperation in 2D involves first-neighbor connections, while cooperation in 1D involves first- and second-neighbor connections. Competition has to be implemented through the AER communication protocol, and it is therefore flexible in terms of connectivity pattern.

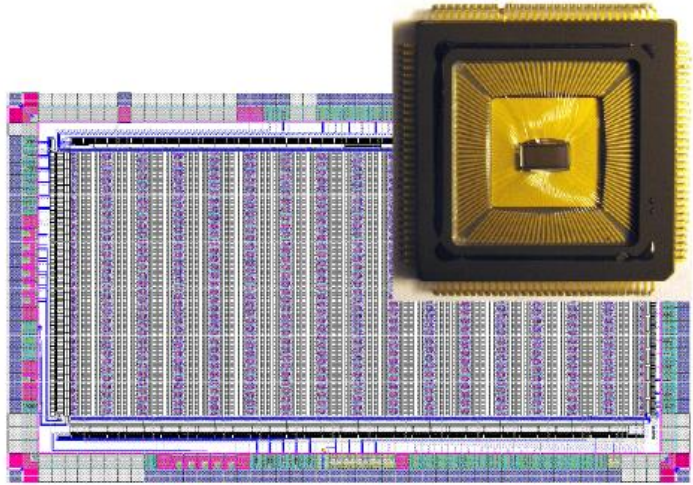

Figure 18. 2DIFWTA chip layout and photograph. The 2DIFWTA chip was implemented using a standard 0.35 µm four-metal CMOS technology and covers an area of about 15 mm².

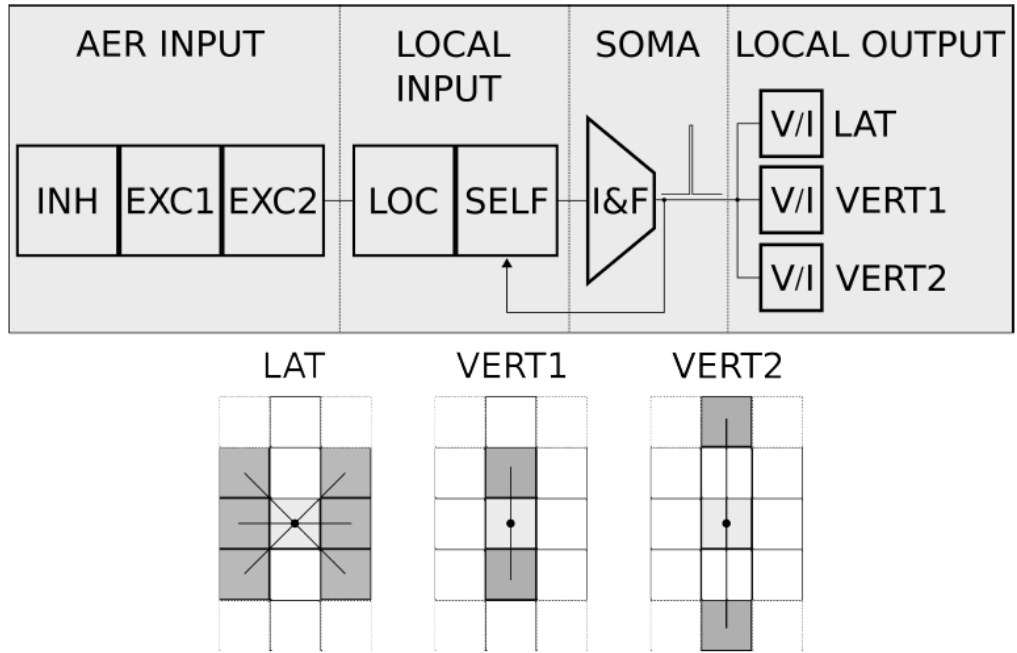

Figure 19. Neuron diagram (top) and local connectivity (bottom). Each neuron comprises the following blocks: AER input, local input, soma and local output. The AER input consists of three AER synaptic circuits: two excitatory and one inhibitory. The local input comprises a diff-pair circuit to integrate all the local excitatory input (see local connectivity diagram below) and an excitatory synapse to implement self-excitation. The soma is an I&F neuron circuit that integrates the sum of all currents generated by the AER input and local input blocks. The local output block generates pulse currents for the local input blocks of the neighbors' neurons. The pattern of recurrent local connectivity is represented in the bottom diagram. Local cooperation of the 2D network is implemented by activating the lateral ("LAT") and vertical to first neighbors ("VERT1") local connections. Several 1D networks with first- and second-neighbor cooperation are implemented by activating the vertical to first neighbors' ("VERT1") and vertical to second neighbors' ("VERT2") recurrent excitatory connections.

## 6.2 Neuron (I&F) model

The circuit diagram of the I&F neuron implemented on the 2DIFWTA chip is shown in Figure 20. It is a compact leaky I&F circuit optimized for power consumption based on the design proposed in [77], that implements spike-frequency adaptation as well as a tunable refractory period, and voltage threshold modulation. The spike frequency adaptation mechanism used in our silicon neuron models the effect of calcium-dependent after-hyperpolarization potassium currents present in biological neurons [78]. Continuous improvements in VLSI technology allow for the fabrication of AER devices containing vast numbers of spiking elements operating in parallel. These devices will be practically realizable only if the spiking neuron circuits have minimal power consumption locally, implement pulse-frequency saturation (refractory periods) for limiting the power consumption globally, and contain spike frequency adaptation mechanisms to reduce communication bandwidth for the transmission of address-events. A detailed description of the circuit operation can be found in [21].

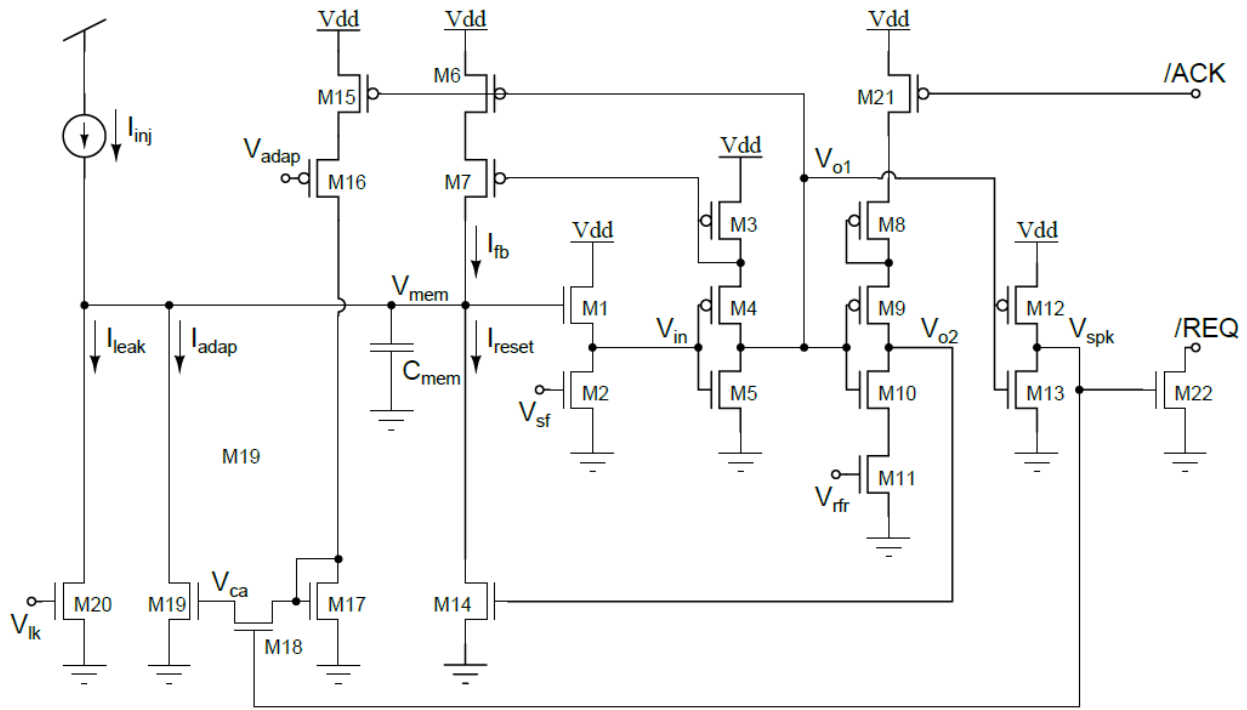

Figure 20. Circuit diagram of an I&F neuron. The input current $I_{in}$ is integrated on to the neuron's membrane capacitor $C_{mem}$ until the spiking threshold is reached. At that point, the output signal $V_{spk}$ goes from zero to the power supply rail, signaling the occurrence of a spike. Then the membrane capacitor is reset to zero, and the input current starts to be integrated again. The leak module implements a current leak on the membrane. The spiking threshold module controls the voltage at which the neuron spikes. The adaptation module subtracts a firing rate dependent on current from the input node. The amplitude of this current increases with each output spike and decreases exponentially with time. The refractory period module sets a maximum firing rate for the neuron. The positive feedback module is activated when the neuron begins to spike and is used to reduce the transition period in which the inverters switch polarity, dramatically reducing power consumption. The circuit's biases ($V_{lk}$, $V_{adap}$, $V_{alk}$, $V_{sf}$, and $V_{rf}$) are all sub-threshold voltages that determine the neuron's properties.

## 6.3 Synapses

The synaptic circuit implemented in this chip is the one proposed by Bartolozzi and Indiveri [67]and referred to as Differential Pair Integrator (DPI) synapse. The primary motivation for choosing this synaptic circuit for the 2DIFWTA chip relates to its linear filtering properties. The two-dimensional structure of this chip implies strong limitation on the space available for synaptic circuits provided the goal of integrating a large number of neurons. The linearity of the synaptic circuit allows multiplexing of different spiking sources, so that a single synaptic circuit can virtually act as multiple synapses. This property is used in the 2DIFWTA chip both for the input AER connections and the local connections, whenever a single time constant can be accepted.

## 6.4 Full neuron and local connectivity

The full neuron comprises several blocks, as depicted in Figure 19. The two AER excitatory synapses labelled "EXC1" and "EXC2" have independent bias settings and are intended for providing external stimuli or implementing arbitrary recurrent connections. The AER inhibitory synapse ("INH") can be used to transmit global inhibition signals in a cooperative-competitive network configuration, but it is also suitable for providing any external and recurrent inhibitory input. The local input block comprises an "LOC" circuit for summing up all contributions from local cooperative (e.g., excitatory) connections and integrate them through a DPI synapse (see Section 6.3). The "SELF" block implements self-excitation. The "I&F" circuit is described in Section 6.2. The local output generates contributions for neighbor neurons (to be integrated in the respective "LOC" block). The connectivity scheme for the recurrent excitation (as depicted in the bottom part of Figure 19) has been designed to provide two kinds of cooperative-competitive networks. The "LAT" and "VERT1" connections can be activated simultaneously to implement local cooperation in a 2D network. Otherwise, the "VERT1" and "VERT2" connections can be activated simultaneously for implementing 32 1D networks of 64 neurons with first- and second-neighbor excitatory recurrent connections. All synaptic circuits included in the "AER INPUT", "LOCAL INPUT", and "LOCAL OUT-PUT" blocks can be independently activated or de-activated, therefore providing full flexibility for the effective connectivity matrix of the 2048 neurons.

## 6.5   Application

Despite the very specific motivation of designing the 2DIFWTA chip for implementing cooperative-competitive networks, its architecture is such that the chip can be treated as a general-purpose, large scale pool of neurons. Therefore, it has been possible to use this chip to address a broad range of research questions, as demonstrated by the number of publications making use of this hardware.

The 2DIFWTA chip was used to evaluate the role of global inhibition in a hardware model of olfactory processing in insects [79]. Function approximation [80] and inference [81] were implemented on neuromorphic hardware by making use of multiple 1D cooperative-competitive networks available on the 2DIFWTA chip. Both first- ("VERT1") and second-neighbor ("VERT2") recurrent excitatory connections were activated for these experiments.

In [82], the chip was used as a large pool of neurons (with no recurrent connections) configured to maximize mismatch in the generation of the action potential in response to a single input pulse, therefore producing a distribution of axonal delays in the range of milliseconds. This result was instrumental in the later implementation of a neuromorphic system for unsupervised learning of simple auditory features [83].

Probably the most advanced application of this chip was presented in [84], where a real-time neuromorphic agent able to perform a context-dependent visual task was demonstrated. The paper presented a systematic method for configuring reliable Finite State Machine (FSM) computation on spiking neural arrays implemented on neuromorphic hardware.

The 2DIFTWA was later used to explore models of auditory perception in crickets [85] and to study latency code processing in the weakly electric fish [86].

## 7   PARCA :Parallel Architecture with Resistive Crosspoint Array

Integration of resistive synaptic devices into crossbar array architecture can efficiently implement the weighted sum or the matrix-vector multiplication (read mode) as well as the update of synapse weight (write mode) in a parallel manner. The parallel architecture with resistive crosspoint array (PARCA) [87] is shown in Figure 21(a), and employs read and write circuits on the periphery.

To compute the weighted sum, the RRAM array is operated in the read mode. Given an input vector $x$, a small read voltage $V_{row,i}$ is applied simultaneously for every non-zero binary bit of $x$ (Figure 21(b)). $V_{row,i}$ is multiplied with the conductance $G_{ij}$ at each cross point, and the weighted sum results in the output current at each column end. As shown in Figure 21(d), the column read circuit integrates this analog current and converts to spikes or digital output values (i.e., current-to-digital converter) with a non-linear activation function (e.g., thresholding), such that the proposed architecture performs the analog computing only in the core of the crossbar array, and the communication between arrays is still in a digital fashion. Compared to conventional memories that require row-by-row read operation, this approach reads the entire RRAM array in parallel, thereby accelerating the weighted sum.

To update the resistive synapse weights, the RRAM array is operated in the write mode, with local programming voltage generated at local row and column peripheries (Figure 21(c) and (d)). This approach can implement the gradient descent or spike-based learning algorithm where the intended synapse weight change is mapped to the conductance value change of RRAM devices. The column write circuit (Figure 21(c)) generates a programming pulse with a duty cycle proportional to column neuron value. The row write circuit (Figure 21(b)) generates a number of pulses proportional to the row neuron value, where the pulse width is fixed and the pulses are evenly distributed across a constant write period to minimize the quantization error. During the write period, the conductance of RRAM cells is changed by the aggregate overlap time of the column write window ($y_j$) and pulses on the row side ($x_i$), which effectively represents $x_i y_j$.

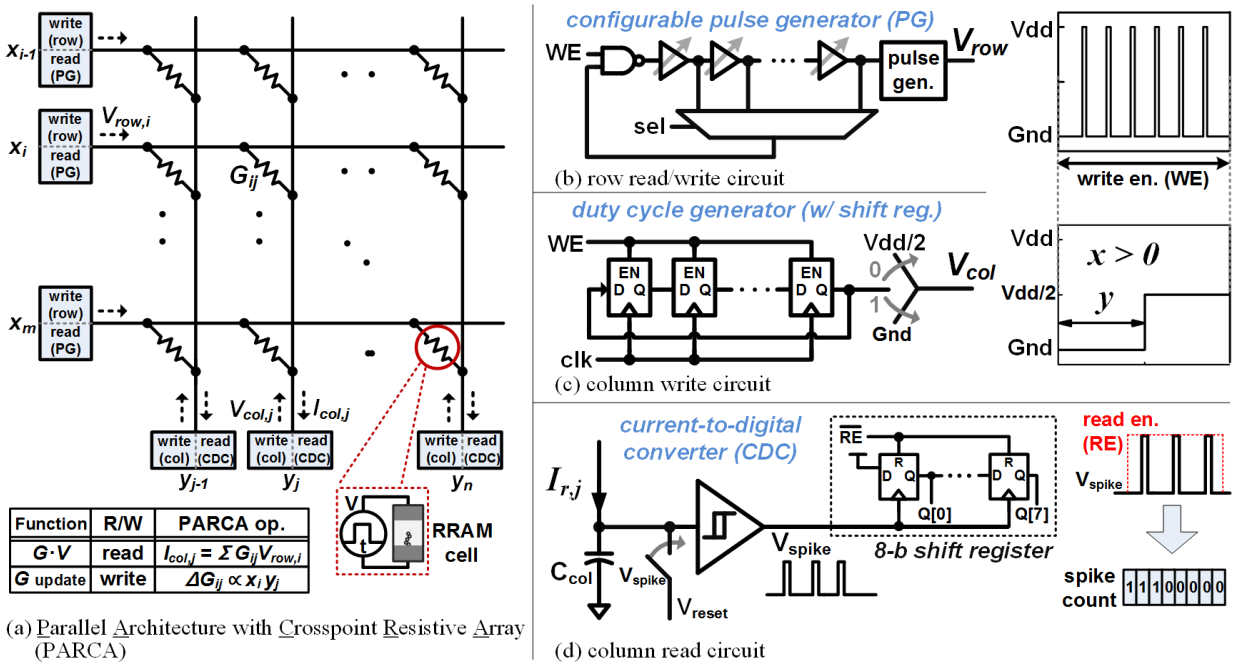

Figure 21. Parallel architecture with resistive crosspoint array (PARCA). (a) Parallel architecture of resistive crosspoint array (PARCA) is shown. (b) Row read/write circuit modulates the number of pulses over time. (c) Column write circuit generates the duty cycle window where write is effective. (d) Column read circuit converts analog output current (weighted sum) into digital values. Adapted from [87].

So far, there have been a few experimental implementations of simple algorithms on small-scale crossbars: 40×40 Ag:a-Si crossbar (loading weights without learning) [88], 12×12 TiO$_x$/Al$_2$O$_3$ crossbar (simple perceptron algorithm with software neurons) [89], IBM's 500×661 PCM 1T1R array (multi-layer perceptron with software neurons) [90], and IBM's 256×256 PCM 1T1R array (with on-chip integrate-and-fire neurons) [91]. Recently, we have designed and fabricated a 64×64 neurosynaptic core with RRAM 1T1R synaptic array and CMOS neuron circuits at the periphery, as shown in Figure 22. The RRAM is monolithically integrated between M4 and M5 in a 130 nm CMOS process. In this design, RRAM is a conventional binary device, suitable for binary neural networks. The CMOS neuron at column peripheries have been time-multiplexed to minimize neuron area and sequentially read out different columns.

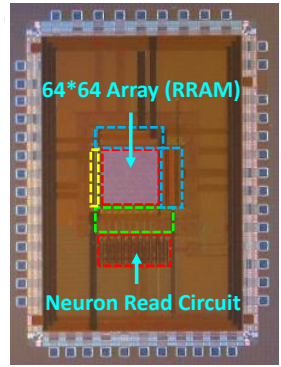

Figure 22. Crossbar macro die.

From the device perspective, today's resistive synaptic devices with multilevel states could emulate the analog weights in the neural network and the crossbar array could efficiently implement the weighted sum and weight update operations. However, non-ideal device effects exist when mapping the weights in the algorithms into the conductance of the devices, including: 1) precision (or number of levels) in the synaptic devices is limited as opposed to the 64-bit floating-point in software; 2) weight update (conductance vs. # pulse) in today's devices is nonlinear and asymmetric (see representative examples of PCMO [92], Ag:a-Si [93] and $TiO_2/TaO_x$ [94] devices in literature); 3) weight on/off ratio is finite as opposed to the infinity in software, as the off-state conductance is not perfectly zero in realistic devices; 4) device variation, including the *spatial* variation from device to device and the *temporal* variation from pulse to pulse, is remarkable; 5) at array-level, the IR drop along interconnect resistance distorts the weighted sum. In order to evaluate the impact of non-ideal effects of synaptic devices and array parasitics on the learning accuracy at the system-level, it is necessary to formulate a methodology to abstract the device's behavioral model and incorporate it into the key operation steps in the algorithm. Pioneering works [95], [96] have been performed to study the non-ideal device effects using sparse coding algorithm as a case study. Sparse coding is an unsupervised learning algorithm for efficient feature extraction and it is bio-plausible: neurons in the visual cortex can form a sparse representation of natural scenes [97]. The device-algorithm co-simulation flow is shown in Figure 23(a). The MNIST handwritten digits dataset [98] is used for benchmark. Figure 23(b) shows the learning accuracy with different precisions by truncation of the bits of the neuron vector (Z) and weight matrix (D) in the algorithm. At least 6-bit D and 4-bit Z are needed for high learning accuracy. This translates to 64 levels of conductance in synaptic devices, which is available in today's devices. However, with a naïve implementation of today's realistic synaptic devices, it only achieves a poor recognition accuracy (~65 %) of the MNIST handwritten digits compared to that with ideal sparse coding algorithm (~97 %), as shown Figure 23(c). The impact of the non-ideal effects is evaluated one by one, and it is summarized in Table 3.

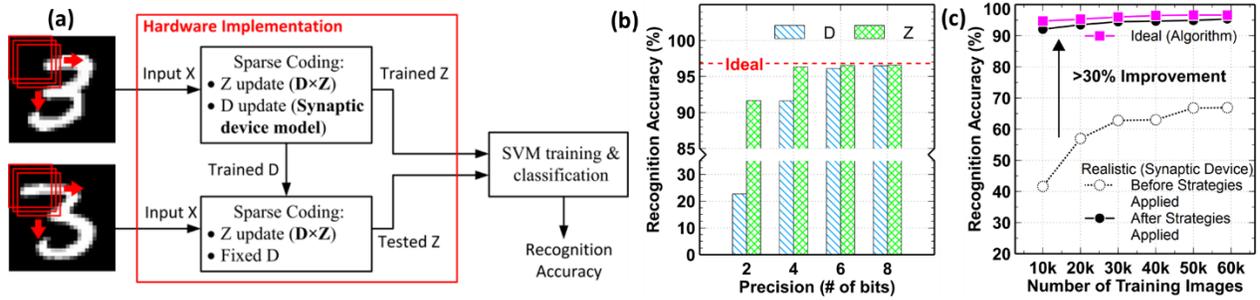

Figure 23. Device-algorithm co-simulation flow (a) Flow of sparse coding with non-ideal effects modeled in the algorithm. (b) Recognition accuracy as a function of precision of bits of the neuron vector (Z) and weight matrix (D). (c) Recognition accuracy with realistic device behaviors and the improvement by the proposed mitigation strategies. Adapted from [95], [96].

To mitigate these non-ideal device effects, strategies at circuit- and architecture-level were proposed [95], [96], and are also summarized in Table 3. These include a dummy column to eliminate the off-state current, smart programming schemes to improve the linearity of weight update, the use of multiple cells as one-weight bit to statistically average out device variations, and relax the wire width for reducing the IR drop. With the proposed strategies applied, the recognition accuracy can increase back to ~95 %. These mitigation strategies are associated with penalty of area, power, and latency, etc.

Table 3. Summary of non-ideal effects and mitigation strategies (Adapted from [95], [96])

| Non-ideal Effects | Impact on Learning Accuracy | Mitigation Strategies |
|---|---|---|
| Limited weight precision | Significant when <32 levels | Need >64 levels by device engineering |
| Nonlinear weight update | 5% drop is nonlinearity is large (when ~1/3 number of pulses is enough to increase the weight from 0 to 90% full) | Smart programming scheme |
| Limited on/off ratio | ~20% drop when on/off ratio <15 | Dummy column and differential read-out |
| Device spatial variation | Not significant even when variation is 30% | Multiple cells as one bit to average out variation by redundancy |
| Device temporal variation | ~20% drop when variation is 30% | |
| IR drop on the array wire | >7% drop is wire width<50nm for 900*300 array | Relax the wire width |

## 8    Transistor-channel based programmable and configurable neural systems

This section discusses the neural system design and potential of transistor-channel modeling. Modeling of biological dynamics in physical implementation results in practical applications, consistent with Mead's original vision (e.g., [4]). One does not require choosing modeling neurobiology *or* building practical applications. Neurobiology seems optimized for energy-efficient computation, something engineers explicitly want for embedded applications. Long-term memory elements enable computation as well as robustness from mismatch. These approaches can become the basis for building models of the human cortex as well as having impact for commercial applications [99].

### 8.1    Transistor-channel models of neural systems

The base components are based on transistor channel models of biological channel populations [100] (Figure 24). The physical principles governing ion flow in biological neurons share similarities to electron flow through MOSFET channels, and exploiting these similarities results in dense circuits that effectively model biological soma behavior. The energy band diagram (source to drain) looking through the channel of the MOSFET is similar to the energy band diagram (inside to outside) looking through a biological channel. Because the similarities between biological and silicon channels are utilized, the voltage difference between the channel resting potentials on the silicon implementation is similar to that in the biological power supplies. The resulting spiking action-potential circuit requires six transistors (Figure 24), which is the same number of transistors and just a few more capacitors (transistor-size capacitors) than the basic integrate-and-fire neuron approach [4].

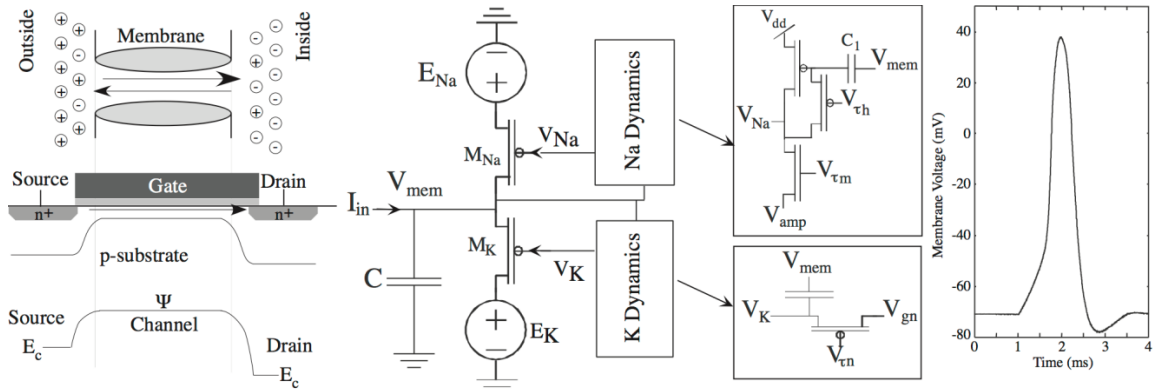

Figure 24. An overview of MOSFET channel modeling of biological channels. This approach is possible given the similar (although not identical) physics between MOSFET and biological channels, both modulated by a gating potential. The physical structure of a biological channel consists of an insulating phospholipid bilayer and a protein that stretches across the barrier. The protein is the channel in this case. The physical structure of a MOSFET consists of polysilicon, silicon dioxide, and doped n-type silicon. A channel is formed between the source and the drain. The band diagram of silicon has a similar shape to the classical model of membrane permeability. This approach yields an updated model for modeling biological channels that also empowers dense MOSFET implementation of these approaches. The primary design constraint is modeling the gating function with other transistor devices; such an approach is shown to model the classic Hodgkin–Huxley squid axon data, resulting in a close model to the event, as well as voltage clamp experiments.

## 8.2    Long-term analog memories: Dense silicon synapse models

Long-term memory enables computation, including physical computation, which encompasses analog and neuromorphic computation. A Floating-Gate (FG) device is employed that can be used to store a weight in a nonvolatile manner, compute a biological excitatory post-synaptic potential (EPSP), and demonstrate biological learning rules (Figure 25) (P. Hasler, Diorio, Minch, & Mead), [102], [103].

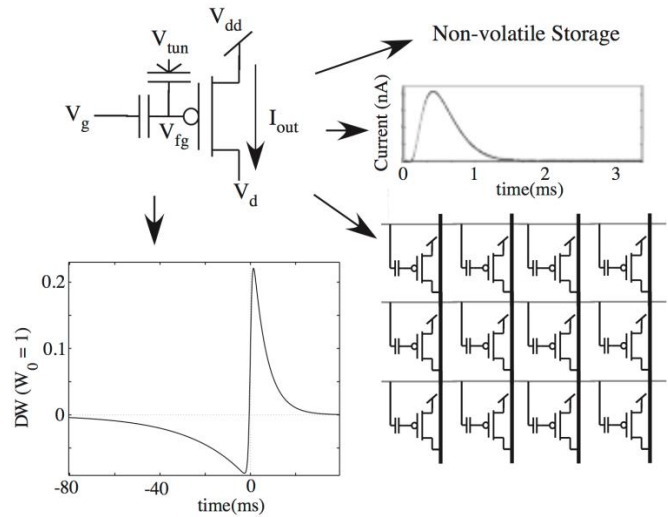

Figure 25. A single transistor synapse device is presented; the architecture uses non-volatile storage, generates biological post-synaptic potential (PSP) outputs, can easily be arrayed in a mesh architecture, and demonstrates biological synapse learning rules, such as long-term potentiation (LTP), long-term depression (LTD), and spike time dependent plasticity (STDP).

Synapses represent the connection between axon signals and the resulting dendrite of a particular neuron. The connection starts as an electrical event arriving into the presynaptic cell, releasing chemicals that reach and modulate the channels at the postsynaptic cell, resulting in a response in the dendritic structure. Figure 25 shows single transistor learning synapse using a triangle waveform modeling the presynaptic computation, a MOSFET transistor modeling the postsynaptic channel behavior, and a floating-gate to model the strength of the resulting connection. A MOSFET transistor in sub-threshold has an exponential relationship between gate voltage and channel current; therefore achieving the resulting gate voltage to get the desired synapse current, which has the shape of a triangle waveform. These learning synapses have storage capabilities to enable them to retain 100s of quantization levels ($7-10$ bits), limited by electron resolution, even for scaled down floating-gate devices (i.e., 10 nm process).

Biological synapses adapt to their environment of event inputs and outputs, where typical programming rules include long-term potentiation (LTP), long-term depression (LTD), and spike-time-dependent plasticity (STDP). A single floating-gate device has enabled both the long-term storage and PSP generation, but also has allowed a family of LTP-, LTD-, and STDP-type learning approaches through the same device [102]. The weight increases when the postsynaptic spikes follow the presynaptic spikes and decreases when the order is reversed. The learning circuitry is again placed at the edges of the array at the end of the rows, included in the soma blocks, therefore not limiting the area of the synaptic matrix/interconnection fabric.

### 8.3    Neuromorphic IC of somas and learning synapses

Figure 26 shows an IC implementation of biological model circuits for synapses, soma (channel model), and input and output spikes efficiently into the system [104]. A mesh architecture (or crossbar, as originally described in [101]) enables the highest synapse array density interconnecting configurable, channel-neuron model components. The soma has the capability of multiple channels, infrastructure for communicating action potential events to other structures, as well as circuits to build local WTA feedback between the soma membrane voltages. This configurable array is a specialized large-scale Field Programmable Analog Array (FPAA) (e.g., [105]). The array of synapses compute through a memory that is similar to an array of EEPROM devices; any off-synapse memory-based scheme will only be more complex and expensive system design. The chip uses AER, with the digital computation synthesized from Verilog using standard digital cells. The synapses were enabled for STDP learning operation and initial learning experiments performed [106].

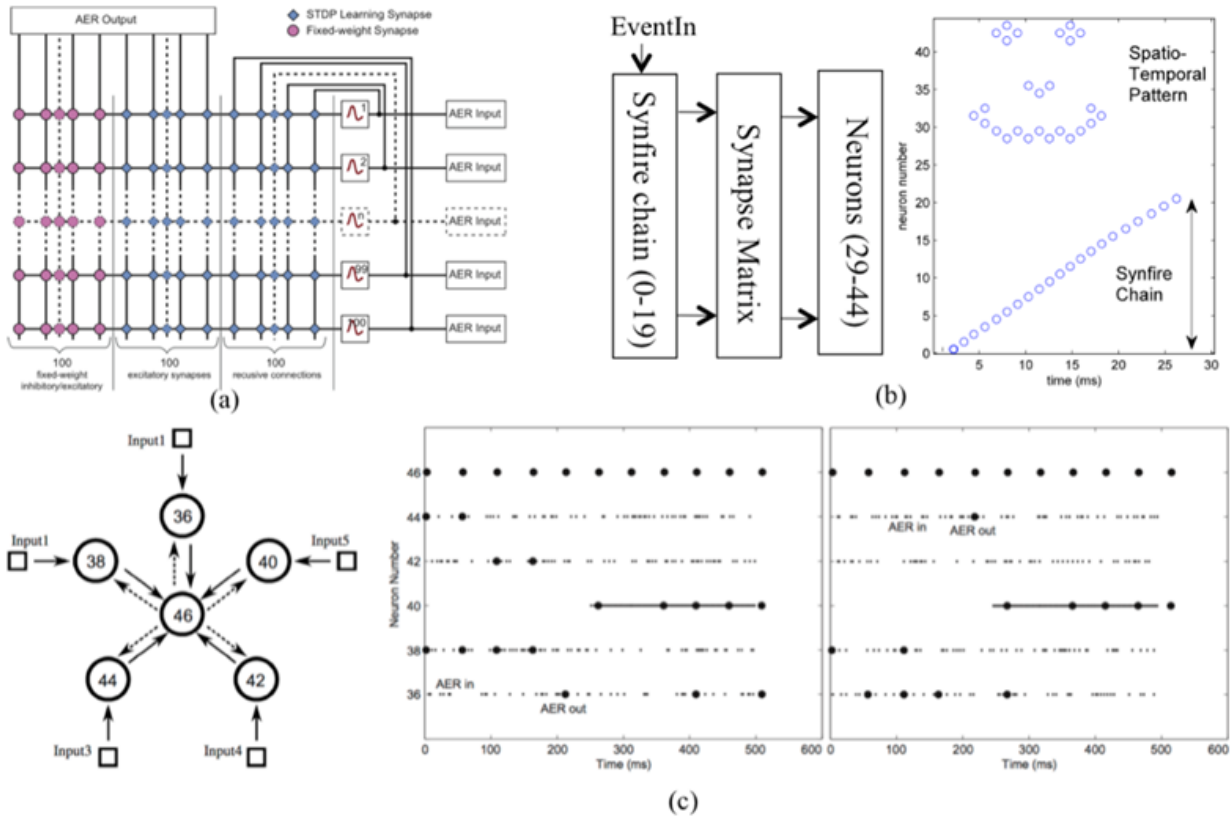

Figure 26. Neuron IC built from transistor channel modeled components. (a) The IC uses a mesh-type structure of synapses, an array of configurable elements to compile the desired soma dynamics, and AER blocks that output action potentials into the synapse fabric as well as AER blocks that that input action potentials from soma blocks, 10,000 STDP excitatory learning enabled synapses connected in recurrent connections, 10,000 STDP excitatory learning enabled synapses connected from AER, and 10,000 programmable (fixed-weight) excitatory or inhibitory synapses connected from AER. (b) Measured synfire neuron chain activating other programmed neurons. (c) Diagram and experimental measurements of a spiking Winner-Take-All (WTA) network drawn as a ring topology. The network is a combination of input neurons on the outside part of the ring, and an interneuron, which provides the negative feedback through inhibitory neurons, for the WTA computation. We label the particular neurons used for this experiment, and all synapses used are programmable synapses, whether excitatory or inhibitory.

Table 4 shows comparison between the best IC synaptic density [[104], [107], [108], [21], [109]]. These results demonstrate the resulting advantage of floating-gate approaches for neuromorphic engineering applications. These techniques will scale down and have relatively similar density to EEPROM density at a given process node [110].

Table 4. Comparison of synapse density and function of working implementations [104]. Synapse density is the synapse area normalized by the square of the process node.

| Chip Built | Process Node (nm) | Die Area (mm²) | # of synapse | Synapse area (µm²) | Syn density | Synapse Storage Resolution and Complexity |
|---|---|---|---|---|---|---|
| GT Neuron1d [104] | 350 | 25 | 30000 | 133 | 1088 | > 10bit, STDP |
| FACETs chip [107] | 180 | 25 | 98304 | 108 | 3338 | 4bit register |
| Stanford STDP [108] | 250 | 10.2 | 21504 | 238 | 3810 | STDP, no storage |
| INI Chip [21] | 800 | 1.6 | 256 | 4495 | 7023 | 1bit w/ learning dynamics |
| ISS + INI Chip [109] | 350 | 68.9 | 16384 | 3200 | 26122 | 2.5bit w/ learning dynamics |
| ROLLS Chip [9] | 180 | 51.4 | 131584 | 252 | 7778 | 1bit w/ learning dynamics |

Figure 26 shows an example of a synfire chain of neurons. These neurons were programmed with some mismatch compensation, although the channel models were not programmed using mismatch compensation. Additional neurons can be used for additional computations, including showing an artificial pattern from spikes.

Figure 26 shows a WTA topology composed of multiple (5) excitatory neurons that all synapse (excitatory synapses) onto a single neuron that provides inhibitory feedback connection to all of the original neuron elements. The strength of the excitatory connection between each excitatory neuron and the center inhibitory neuron is programmed identical for all excitatory neurons. The strength of the inhibitory connection between the center inhibitory neuron and each excitatory neuron is programmed to identical values, although the strength and time duration of the inhibition time was stronger than the excitatory inputs.

This IC was used to demonstrate optimal path planning based on realistic neurons [111]. Others (e.g., [112]) have built on these approaches. Figure 27 shows how a maze environment is modeled using the neuron IC. A wavefront was initiated at the goal (Node 77) and propagated throughout the neuron grid. Figure 27 shows a raster plot of the solution nodes. Neuron 77 causes neuron 76 to fire, which causes neuron 75 to fire, and so forth. Theoretical *and* experimental results show 100% correct and optimal performance for a large number of randomized maze environment scenarios. The neuron structure allows one to develop sophisticated graphs with varied edge weights between nodes of the grid. Neuron IC's planner analytical time and space complexity metrics were verified against experimental data.

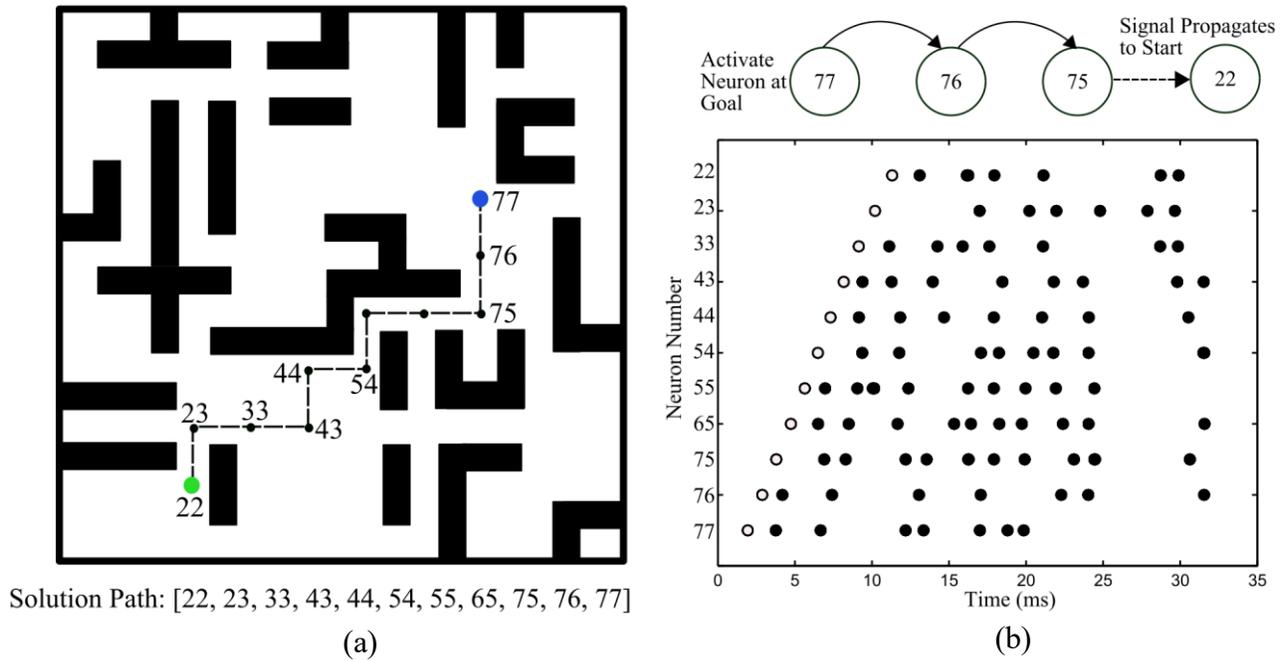

Solution Path: [22, 23, 33, 43, 44, 54, 55, 65, 75, 76, 77]

(a)            (b)

Figure 27. Optimal path planning originally demonstrated on the IC in Figure 26. (a) Grid environment example that the neuron IC solved. The device is located at Node 22 and the goal is at Node 77. A wavefront in initiated at the goal (Node 77) and propagated throughout the neuron grid. (b) Raster plot of the solution nodes. Neuron 77 causes neuron 76 to fire, which caused neuron 75 to fire, etc.

### 8.4 Reconfigurable neuromorphic ICs of dendrites, learning synapses, and somas

Figure 28 shows a further neuromorphic IC design now including dendritic computation, as well as FPAA re-configurability, into the resulting architecture. In many modeling and implementation approaches, the dendrite is approximated to be a wire, greatly simplifying the resulting network and enabling a system that is tractable by a range of computational principles. Using channel model approaches, one can successfully build dense dendritic compartments and configurable structures [113] that compare well to classical models of dendrites [114].

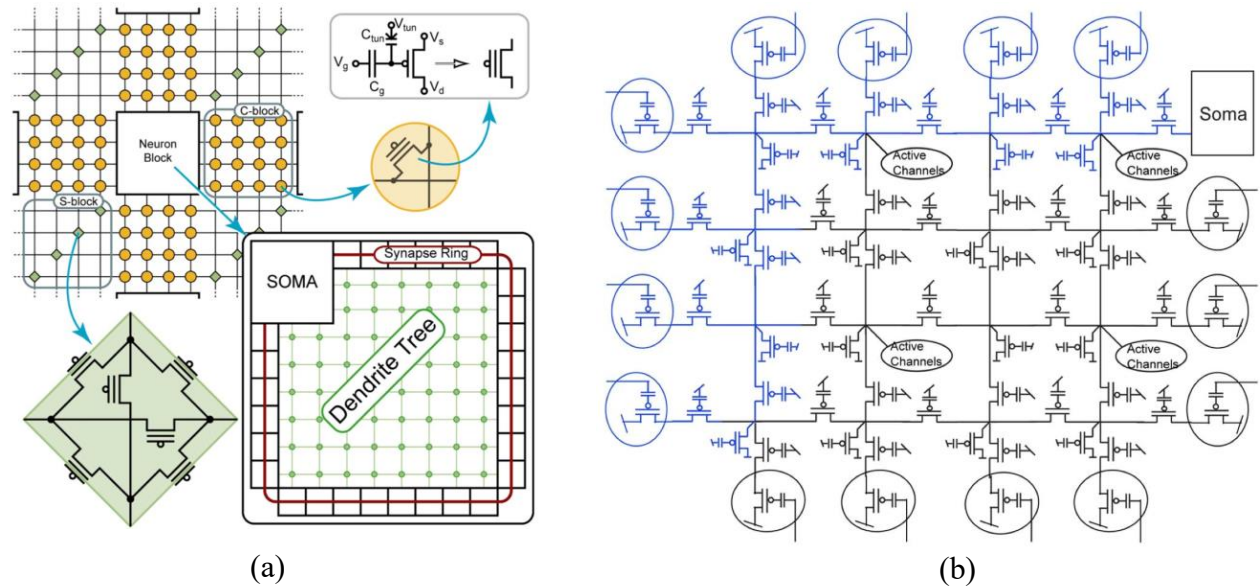

(a)                  (b)

Figure 28. Neuron IC embedded in a typical FPAA/FPGA routing fabric. (a) The Neuron I/O interface with the routing at the C-Blocks through programmable floating-gate switches. The tracks are segmented for allowing faster event transmission and maximizing utilization. The tracks are routed at the S-Blocks, where each node consists of six switches. The neuron cell has synaptic inputs, programmable dendrites with active channels that aggregate inputs into the soma block. (b) Depiction of the neuron cell structure, with arbitrary dendritic structure capability. Dendrites are also interspersed with active channels to model the nonlinear behavior observed in biology. In blue, we show an 8-tap dendritic line programmed on the neuron.

Figure 28 shows the architecture from a large-scale IC enabling dendritic modeling in its computation. Adding dendritic elements changes the neural architecture away from a neat crossbar device alignment. Utilizing concepts from FPAA devices (e.g., [105]), a single neuron is implemented as a single component in the routing. Events can be routed to and from the resulting block as typical in FPAA devices. This approach allows for sparse as well as densely connected neuron infrastructures. The dendritic tree has active and passive channels as well as a reconfigurable soma described in the last subsection. The architecture allows high synapse density in a working neural array; learning algorithms can be localized to a given block. Signal lines can be routed to Address-Event senders and receivers (one-dimensional) to interface with off-chip devices; most (discussed further in [99]) neurons have only local connections.

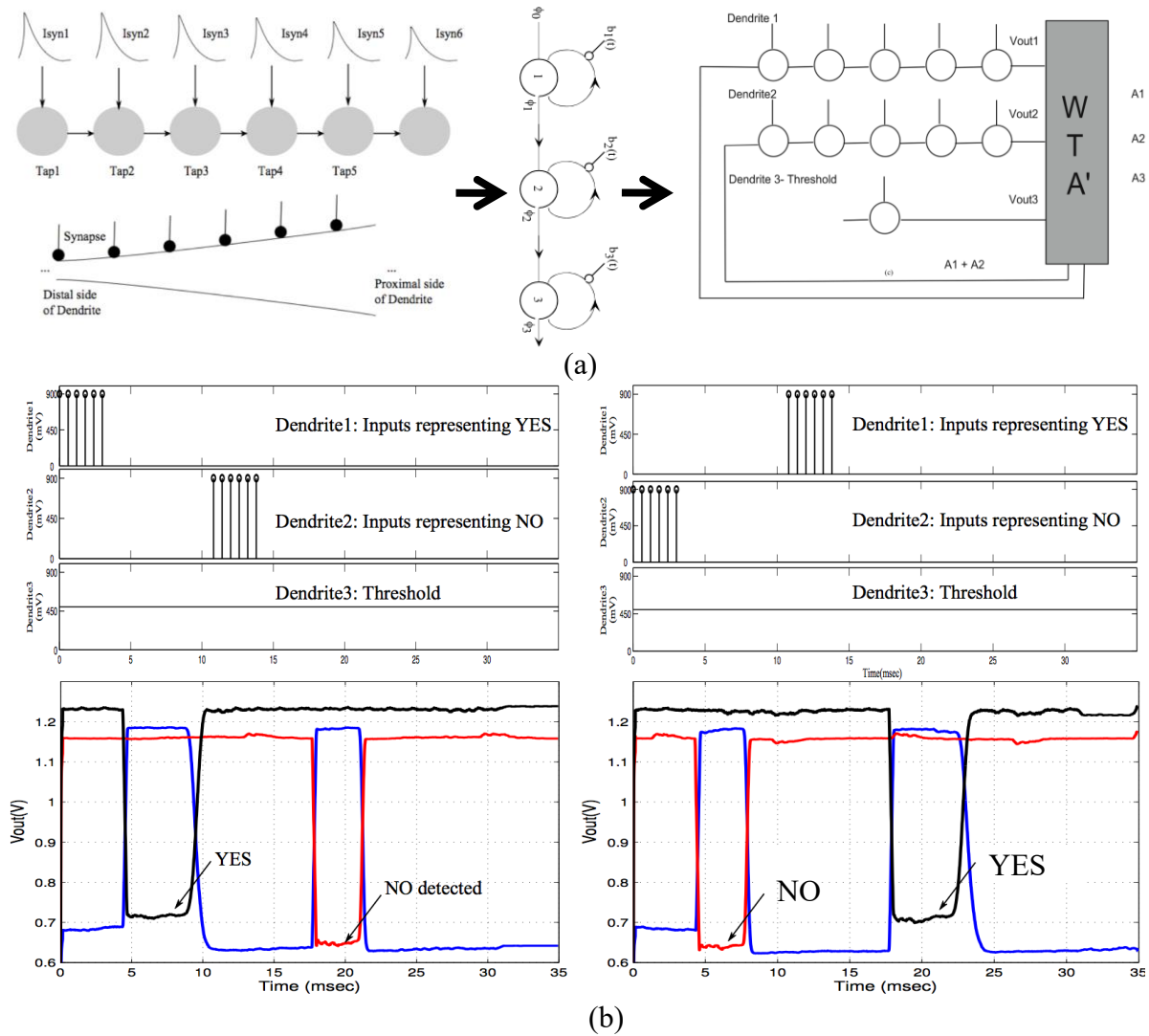

Figure 29. Dendritic computation using the framework in Figure 28. (a) Dendritic trees with increasing diameters have similar function to Hidden-Markov Model (HMM) trees used for acoustic classification. A YES-NO HMM classifier was built using dendritic lines to model linear HMM trees with a WTA terminating block. (b) Experimental demonstration of temporal classification using dendrite enabled neurons to perform temporal classification (word spotting) similar to HMM classifier techniques. In the first case, YES is detected and then NO is detected, whereas in the second case, NO is detected and then YES is detected.

The resulting computation from dendritic elements is often debated, and in most computational models is ignored because of the huge increase in computational complexity. The computational efficiency of dendritic computation (Figure 29) has been demonstrated to be 1000s of times more efficient than that of most analog signal processing algorithms [99], [115]. Dendritic structures can perform word spotting, roughly approximating Hidden-Markov Model (HMM) classifier structures [115]. This computation is critical in a number of engineering applications and typically is seen as too expensive to directly use. With as energy constrained as the cortex would be, as well as the need in embedded electronics for energy-efficient computation, these opportunities cannot be ignored.

### 8.5 Potential directions of channel model computation

Carver Mead hypothesized that physical (e.g., analog) computation would be at least 1000× lower energy compared to digital computation based on transistor counting arguments, an argument experimentally shown multiple times (e.g., [105]). The potential opportunity has started serious discussions of the capability of physical computing (e.g., [116]) building the computing framework, which up to now, has only been developed for digital computing.

The opportunities for neuromorphic computing to impact applications opens up another level of energy-efficient computing. Figure 30 shows the possible opportunities in computing enabling both analog as well as neuromorphic computing. Computation, classification and learning systems that do not show sensor-to-output result capability give a myopic view of their results. These approaches enable opportunities towards building cortical structures [99], [117] as well as enabling many new waves of computing opportunities.

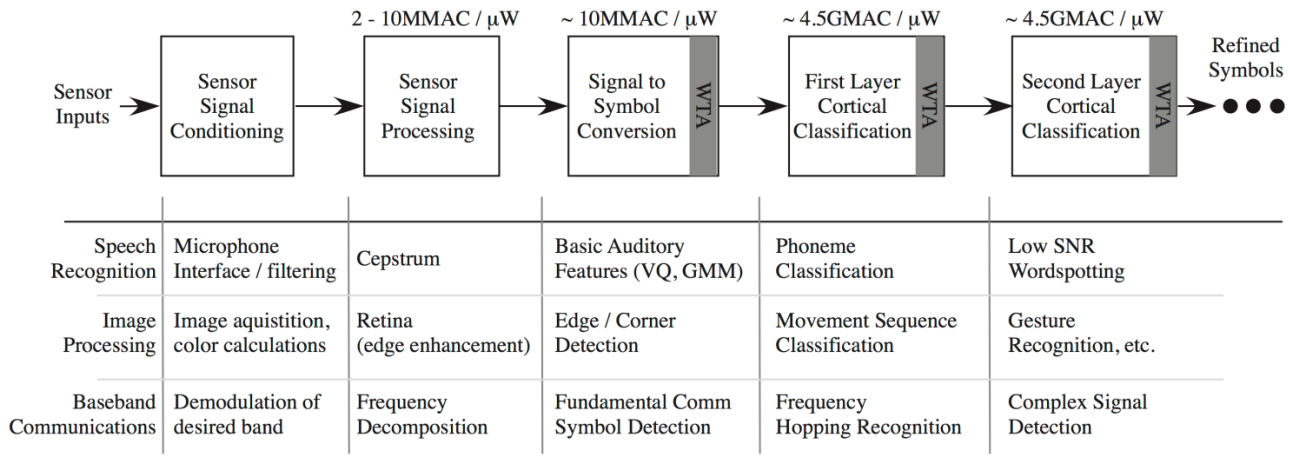

Figure 30. Typical signal processing chain using configurable analog approaches and neural-based classifiers.

# 9    Other neural emulators

## 9.1    Neurogrid

Neurogrid is a mixed-mode multichip system primarily used for large-scale neural simulations and visualization [118].The neuron circuits used in Neurogrid are closely correlated to the physical characteristics of neurons in the brain. It models the soma, dendritic trees, synapses, and axonal arbors. It consists of 16 neurocores/chips each with 65k neurons (totalling 1 M neurons) implemented in sub-threshold analog circuits. A single neurocore is fabricated on an 11.9 mm × 13.9 mm die. A board of 16 neurocores is of size 6.5" × 7.5" and the complete board consumes roughly 3W of power (a single neurocore consumes ~150mW).

## 9.2    TrueNorth

IBMs TrueNorth neuromorphic chip consists of 1 million digital neurons capable of various spiking behaviours [119]. Each die holds 4096 cores, each core holding 256 digital neurons and 256 synapses per neuron. A single die consumes 72 mW of power. They have developed a board (NS16e) comprised of 16 TrueNorth chips, consuming 1W of power at 1KHz speed, making it ideal for energy-efficient applications. Although digital in its implementation, low-power consumption is due to fabrication in an aggressive, state-of-the-art 28 nm technology process.

## 9.3    SpiNNaker

SpiNNaker [120], another digital neuromorphic neural array, was designed for scalability and energy-efficiency by incorporating brain-inspired communication methods. It can be used for simulating large neural networks and performing event-based processing for other applications. Each node is comprised of 18 ARM968 processor cores, each with 32 Kbytes of local instruction memory, 64 Kbytes of local data memory, packet router, and supporting circuitry. A single node consists of 16,000 digital neurons consuming 1W of power per node. There exists two SpiNNaker circuit boards, the smaller being a 4-node (64,000 neurons) board and the largest being a 48-node board (768,000 neurons). The 48-node board consumes 80W of power.

# 10    Discussion

The various state-of-the-art neural simulators described in this paper reveal the evolution and advancement of neuromorphic systems and the neuromorphic engineering field. The promising specifications and applications of these systems advance technology in a manner that further closes the gap between the computational capabilities and efficiency of the human brain and engineered systems. Moreover, it should be clearly noted that the aim of this paper is to merely assess each neural simulator rather than deem one inferior to another. However, depending on one's objectives and specifications of their system, one neural simulator may be more suitable than another. To briefly summarize the key points of each system, the integrate-and-fire array transceiver (IFAT) is a mixed-signal CMOS neural array which was designed to achieve low power consumption and high neuron density (more neurons per $mm^2$). Deep South is a cortex simulator implemented on an FPGA board, and is capable of simulating 20 million to 2.6 billion LIF neurons in real time; further, running 5 times slower, it can scale up to 100 million to 12.8 billion LIF neurons simulations. In addition, it comes with the PyNN programming interface that will enable the rapid modeling of different topologies and configurations using the cortex simulator. The BrainScaleS neuromorphic system is wafer-scale integration of high-speed and continuous time analog neurons and synapses. Wafer module integrates an uncut silicon wafer with 384 neuromorphic chips. It provides programmable analog parameters to calibrate neurons according to models from computational neuroscience. BrainScaleS is faster and it allows to model processes like learning and development in seconds instead of hours. Dynap-SEL is a novel mixed-signal multi-core neuromorphic processor that combines the advantages of analog computation and digital asynchronous communication and routing. It takes the benefits of memory-optimized routing, which makes the system highly scalable [51]. Resources from different chips can be easily combined and merged. 2DIFWTA chip is an implementation of cooperative-competitive networks, and consists of a group of interacting neurons that compete with each other in response to an input stimulus. The neurons with the highest response suppress all other neurons to win the competition. The cooperative and competitive networks give it power to solve complex nonlinear operations. PARCA is a parallel architecture with resistive crosspoint array. Integration of the resistive synaptic devices into crossbar array architecture can efficiently implement the weighted sum or the matrix-vector multiplication (read mode) as well as the update of synapse weight (write mode) in a parallel manner. Transistor-channel based programmable and configurable neural systems is a neuron IC built from transistor channel modeled components. The IC uses a mesh-type structure of synapses, an array of configurable elements to compile the desired soma dynamics, and AER blocks that output action potentials into the synapse fabric, as well as AER blocks that that input action potentials from soma blocks. Furthermore, this neuromorphic IC includes dendritic computation, as well as FPAA re-configurability. The computational efficiency of dendrite computation has been demonstrated to be 1000s of times more efficient than that of most analog signal processing algorithms.

A comparison of all these simulators is depicted in Table 5.

Each of these neurosimulators have advantageous properties that can be utilized depending on one's objective and constraints. Although many of these systems model the dynamics of biological neurons, Neurogrid and the Transistor-channel based systems more closely resemble biological neurons in terms of their physical properties. This sub-threshold, analog representation of neurons is closely related to Carver Mead's original vision. However, many of these systems presented are still capable of emulating many of the prominent neuron spiking behaviors aside from integrate-and-fire (e.g., IFAT, BrainScaleS, Transistor-channel, TrueNorth, and SpiNNaker). DeepSouth, TrueNorth, and SpiNNaker take a digital approach with their neuron design. Although this may be a more robust approach, such a digital design may consume more area and power. TrueNorth has low-power consumption even while using digital technology, but this is due to the state-of-the-art 28 nm feature-size technology that it uses. If one were constrained by power consumption, the IFAT, PARCA, or Transistor-channel based systems would be ideal as these designs were specifically optimized for low-power consumption. However, such low-power design approaches may sacrifice speed and/or accuracy. BrainScaleS and HiAER-IFAT are ideal for high-speed applications as these systems were optimized for high-speed even when simulating large networks. BrainScaleS operates at 10,000× biological real-time speed. HiAER-IFAT uses a novel hierarchical AER design that allows for efficient transmitting and receiving of events over long distances. If one is seeking a system that consists of a well-developed, user-friendly software interface for programming and visualization of the network, the Neurogrid, SpiNNaker, DeepSouth, or TrueNorth might be the appropriate choice. Finally, one may select a system based on a very specific application. For example, for a winner-take-all application (i.e., visual saliency application), one may take advantage of the 2DIFWTA system. If one is interested in on-chip learning, the Dynap-SEL may be ideal. The Dynap-SEL consists of circuitry designed for on-chip learning and further has proven useful as a general-purpose neuron array. This chip was designed in state-of-the-art 28 nm FDSOI technology that further makes it more advantageous with respect to neuron density and low-power consumption.

Overall, each of these systems presents advantages and disadvantages that should not be used to deem one inferior to another. Each of these systems presents a unique capability that may contribute to the growth of the neuromorphic engineering field. These contributions collaboratively move us closer to designing the most energy-efficient, highly dense neural simulator that closely mimics the biological counterpart it is emulating.

Table 5. Comparison of event-based neural processors

| Chip Name | Technology | Process (nm) | Neurons Type | #Neurons | #Synapse | Area per Neuron | #Energy per Event |
|---|---|---|---|---|---|---|---|
| **MNIFAT** | Analog | 0.5 µm | LIF/ M-N | 6120 | - | 1495 $\mu m^2$ | 360 pJ |
| **Deep South** | Digital | - | LIF | 200K | - | | - |
| **Dynap-SEL** | Mixed Signal | 28nm FDSOI | I&F | 1088 | 78080 | | 2.8pJ |
| **BrainScaleS** | Analog Mixed Signal | 180nm | Adaptive Exp IF | 512 | 100 K | 1500 $\mu m^2$ | 100 pJ |
| **2DIFWTA** | Analog | 0.35µm | I&F | 2048 | 28672 | | - |
| **PARCA** | Analog | 130nm | Programmable | - | - | | - |
| **HiAER-IFAT board with 4 chips** | Analog | 90nm | I&F | 256k | 256M | 140 $\mu m^2$ | 22pJ |
| **Transistor-Channel** | Analog | 10nm | Floating Gate MOSFET | - | 10000 | | - |
| **Neurogrid** | Analog | 180nn | Adaptive Quad IF | 65K | 100 M | 1800 $\mu m^2$ | 31.2pJ |
| **TrueNorth** | Digital | 28nm | Adaptive Exp IF | 1M | 256 M | 3325 $\mu m^2$ | 45pJ |
| **SpiNNaker** | Digital | 130nm | Programmable | 16K | 16 M | - | 43nJ |

## 11   Acknowledgement

BrainScaleS work received funding from the European Union Seventh Framework Programme ([FP7/2007-2013]) under grant agreement numbers 604102 (HBP), 269921 (BrainScaleS), and 243914 (Brain-i-Nets), the Horizon 2020 Framework Programme ([H2020/2014-2020]) under grant agreement 720270 (HBP), as well as from the Manfred Stäark Foundation. HiFAT work received funding from NSF CCF-1317407, ONR MURI N00014-13-1-0205, and Intel Corporation. The work is also supported by Pratiksha Trust Funding (Pratiksha-YI/2017-8512), IISc and the Inspire grant (DST/INSPIRE/04/2016/000216) from the Department Of Science & Technology, India.